\documentclass[letterpaper, 10 pt, journal, twoside]{IEEEtran}
\usepackage{cite}

\usepackage{amsmath} 
\usepackage{amssymb}  
\usepackage{color}
  
\ifx\pdfoutput\undefined
	\usepackage{epsfig} 
  	\graphicspath{{Figures/}}
    \DeclareGraphicsExtensions{.eps} 
\else
	\usepackage[pdftex]{graphicx}
	\graphicspath{{Figures/}}
    \DeclareGraphicsExtensions{.pdf} 
\fi
  \usepackage{subfigure}

\usepackage[pscoord]{eso-pic}
\newcommand{\placetextbox}[3]{
\setbox0=\hbox{#3}
\AddToShipoutPictureFG{ \put(\LenToUnit{#1\paperwidth},\LenToUnit{#2\paperheight}){\vtop{{\null}\makebox[0pt][c]{#3}}}}
}
\placetextbox{.5}{0.05}{~\copyright IEEE All rights reserved.Personal use of this material is permitted.  Permission from IEEE must be obtained for all other uses.} \placetextbox{.35}{0.035}{IEEE-RAL-RoboSoft-2021, April 12-16. Virtual Conference hosted by Yale, USA.}

\begin{document}
%
\title{Passive Flow Control for Series Inflatable Actuators: Application on a Wearable Soft-Robot for Posture Assistance}
%
%
%

\author{{Diego} {Paez-Granados}$^{\dagger 1}$, Takehiro Yamamoto$^1$, Hideki Kadone$^2$, and  Kenji Suzuki$^3$%
\thanks{Manuscript received: October, 24, 2020; Revised February, 8, 2021; Accepted February, 28, 2021.}
\thanks{This paper was recommended for publication by Editor Cecilia Laschi upon evaluation of the Associate Editor and Reviewers' comments.
This work was supported by Japanese government MEXT through grant-in-aid: 17H01251.} 
\thanks{$^\dagger$  is the corresponding author.}
\thanks{$^1$ {D.} {Paez-Granados} and  T. Yamamoto are with the Artificial Intelligence Laboratory, Department of Intelligent Interaction Technologies, University of Tsukuba, 1-1-1 Tennodai, Tsukuba 305-8573, Japan.
        {\tt\footnotesize dfpg@ieee.org} {\tt\footnotesize takehiro@ai.iit.tsukuba.ac.jp} }
\thanks{$^3$H. Kadone is with the Center for Innovative Medicine and Engineering, University of Tsukuba Hospital, Japan. 
       {\tt\footnotesize kadone@md.tsukuba.ac.jp}}  
\thanks{$^4$ K. Suzuki is with the Faculty of Engineering and Center for Cybernics Research, University of Tsukuba, Japan. 
        {\tt\footnotesize kenji@ieee.org}  }
\thanks{Digital Object Identifier (DOI): see top of this page.}
}
%
%

\markboth{IEEE Robotics and Automation Letters. Preprint Version. Accepted February, 2021}
{Paez-Granados \MakeLowercase{\textit{et al.}}: Passive Flow Control} 

%



\maketitle

\begin{abstract}
This paper presents a passive control method for multiple degrees of freedom in a soft pneumatic robot through the combination of flow resistor tubes with series inflatable actuators. We designed and developed these 3D printed resistors based on the pressure drop principle of multiple capillary orifices, which allows a passive control of its sequential activation from a single source of pressure. Our design fits in standard tube connectors, making it easy to adopt it on any other type of actuator with pneumatic inlets. We present its characterization of pressure drop and evaluation of the activation sequence for series and parallel circuits of actuators.
Moreover, we present an application for the assistance of postural transition from lying to sitting. We embedded it in a wearable garment robot-suit designed for infants with cerebral palsy. Then, we performed the test with a dummy baby for emulating the upper-body motion control. The results show a sequential motion control of the sitting and lying transitions validating the proposed system for flow control and its application on the robot-suit.
\end{abstract}

\begin{IEEEkeywords}
Soft Robot Materials and Design, Soft robot Applications, Soft Sensors and Actuators, Physically Assistive Devices, Medical Robots and Systems
\end{IEEEkeywords}

\IEEEpeerreviewmaketitle

\section{INTRODUCTION} \label{sec:intro}
\IEEEPARstart{A}{ssistive} robotics is a growing field addressing the need for support for daily activities in response to the decreasing human personnel available in some countries. Furthermore, robotic support in repetitive tasks is highly valued for allowing humans to focus on more challenging and creative tasks. In cases of direct physical interaction with humans, soft-robots present an attractive and encouraging path as the usage of soft-materials increases the compliance to human touch, herewith, reducing the risk of injuries due to impact and making the robots more friendly to the touch.

Within vulnerable populations requiring constant support, children with severe motor disabilities should be prioritized as they are usually bedridden during infancy \cite{Nakada1996}. According to GMFCS (Gross Motor Function Classification System), infants with a level V of cerebral palsy are unable to maintain anti-gravity posture of head and trunk in supine or sitting position by themselves \cite{Palisano1997}, therefore, needing assistance for daily activities and posture transition. Moreover, the progression of cognitive development, sensory development, and motor control are dependent on experiencing the world from different postures \cite{Saavedra2019}.
Equally important, scoliosis, bedsore and joint contractures could be caused by keeping a constant lying posture. Such problems equally affect older adults, people with hemiplegia, and any other condition that forces a bedridden life. 

Mitigation strategies include interventions of posture transition training, and 24-hour posture management \cite{Green2004}. However, in practice, this implies intense nursing care at the base with additional attention to posture, causing hard physical challenges for caregivers. Therefore, there is a need for robotic systems providing posture assistance.

\begin{figure}[!t]
	\centering
	\subfigure[Flow resistors embedded in series inflatable actuators.]{\includegraphics[width=3.7cm]{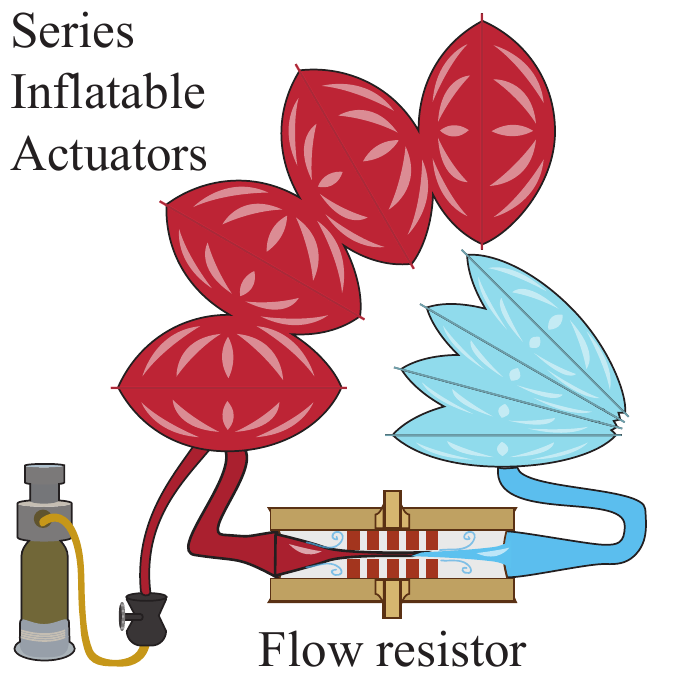}
	\label{fig:image}}
	\hfil 
	\subfigure[Soft-robot suit implemented on an infant dummy.]{\includegraphics[width=4cm]{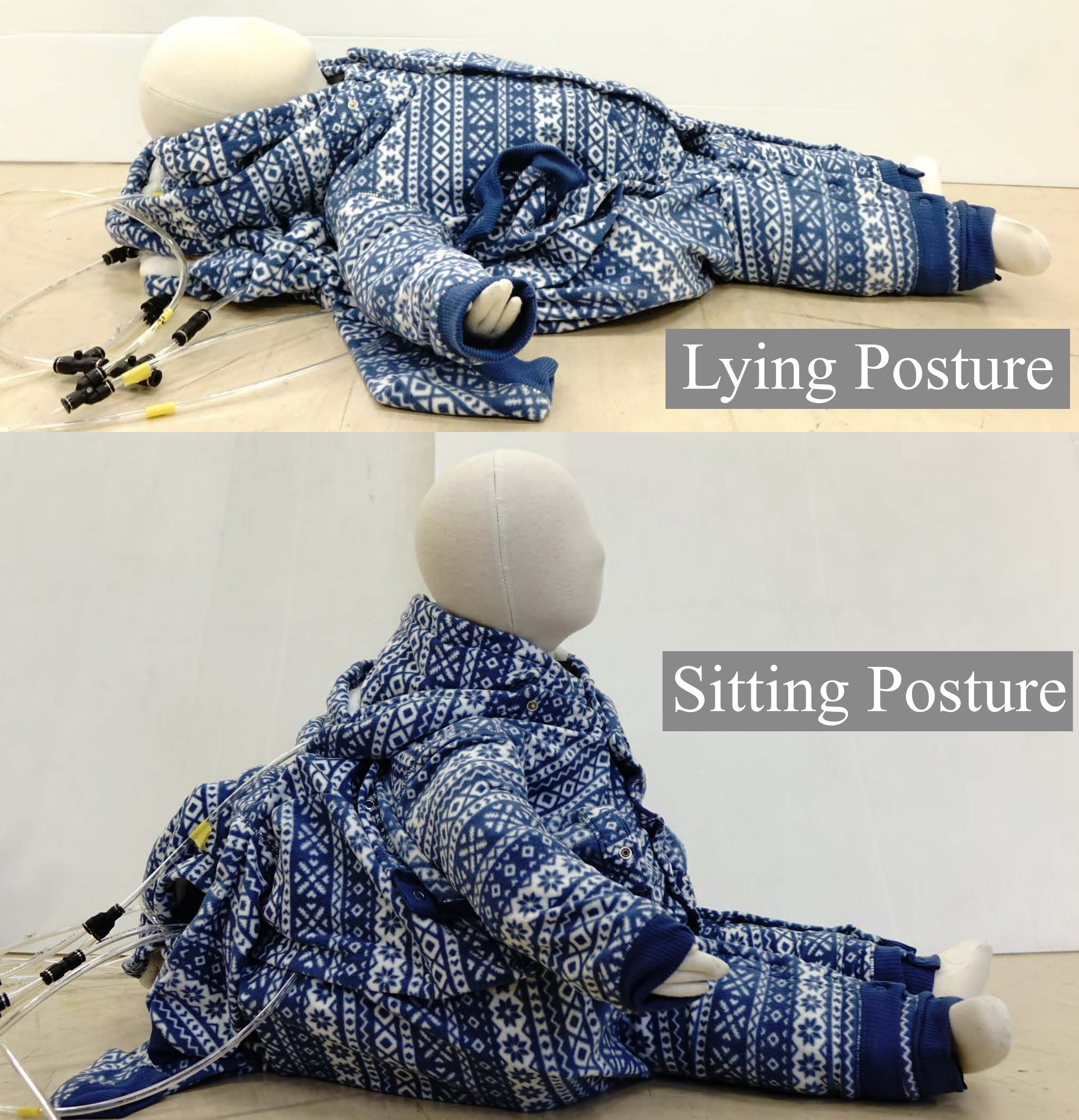}\label{fig:exoskeleton}} 
	\caption{Integrated series inflatable actuators for an electronics-free sequential activation through passive flow resistors, requiring a single on/off bypass for generating the successive activation of each chamber.
	\label{fig:intro} }
\end{figure}

When considering the fragility of the body structure accompanied with joint rigidity due to joint contraction and involuntary muscle contractions, a wearable soft-robot offers a good opportunity to balance support and safety while attaining enough freedom for experiencing the world and improving motor control, something yet not available in any system, especially for children unable to get up from bed by themselves.

\begin{figure}[!t]
      \centering
	    \includegraphics[width=8.4cm]{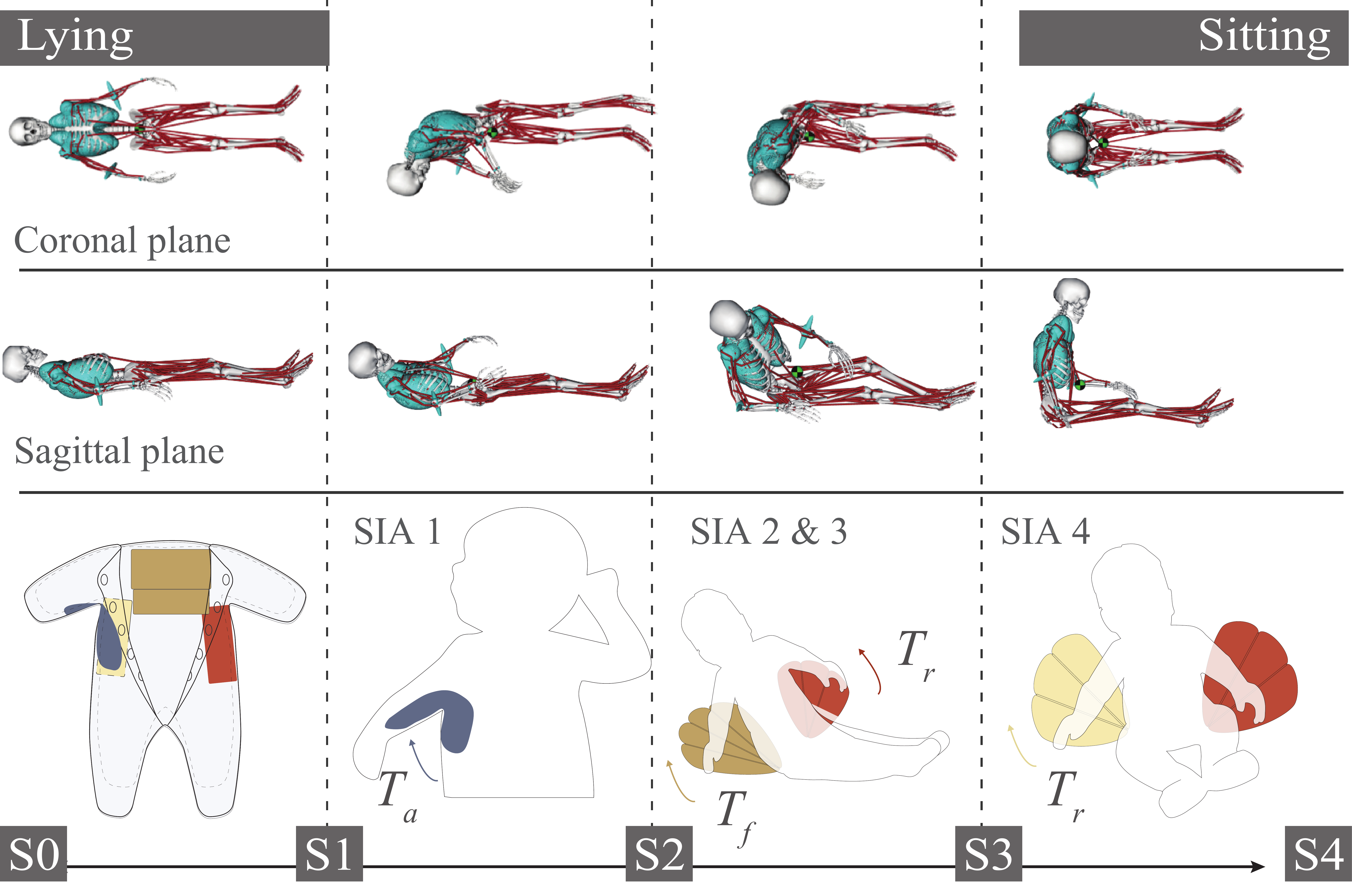}
      \caption{Proposed postural transition assisted through the method of elbow supported mid-phase. On the bottom part, we present the mimicked proposed transition through an embedded set of series inflatable actuators (SIA) as a way of assisting the transition with a soft robot-suit. \label{fig:transitions}}
   \end{figure}

We propose postural transition assistance through a garment embedded with soft inflatable actuators at four limb support locations for a 3-step transition (as depicted in Fig. \ref{fig:transitions}). This method mimics the natural body transitions of a healthy individual (validated in \cite{Yamamoto2015}), thus, motivating natural control of their limbs for potential training. 
Moreover, we propose a passive system that ensures the sequential activation of the multiple degrees of freedom by a single activation valve (as depicted in Fig. \ref{fig:SIA_schematic}, herewith, achieving a flexible system that procures different sequenced motions for the unique needs of the end-users and their caregivers.

The goal of this work was to provide a simple method of controlling the sequential activation of multiple series inflatable actuators (SIA), making it applicable to numerous systems, such as soft-robots motion control \cite{Gilbertson2017}, bending robots deformation and activation  \cite{Futran2018}, and locomotion of inflatable robots \cite{Booth2018}.

\subsection{Contributions}
This work contributes to the field of soft-robotics through:
\begin{enumerate}
    \item A new design and evaluation of a soft robot suit with inflatable actuators for the postural transition from lying to sitting by multiple actuators with multi-chambers bladders.
    \item A design and characterization of flow resistors with variable pressure drop through a fully passive method (unmovable components) introducing multi-orifice plates through the principle of pressure drop in multi-capillary orifices. Consequently, controlling the timed activation of the series inflatable actuators by a single on/off bypass valve.
\end{enumerate}

	
\subsection{Related Work}
Elastic inflatable actuators (EIA) are a possible replacement of software-based compliant and adaptive control \cite{Yang2011}, although a disadvantage exists in the reduced motion control precision, a hardware-based intelligence embedded through engineering design of flexible structures has potential in rehabilitation \cite{Belforte2014, Sridar2018}, robot locomotion \cite{Gorissen2019} and even humanoid robotics \cite{Best2015}. A review of EIA applications and control is detailed in \cite{Gorissen2017}.
For controlling pneumatic circuits with electronics-free methods several works have shown micro-fluids valves that effectively act as logic gates \cite{Rhee2009}. Similar principles have demonstrated pneumatic applications through soft-membranes as recently presented in \cite{Rothemund2018, Preston2019}, with a set of logic gates circuits effectively achieving complex tasks in soft robotics applications, such as locomotion, grasping, and arm motion control. As well, the work in \cite{Xu2020} presented a by-pass valve for achieving a NOT logic gate and subsequent extension to multiple logic circuits for developing pneumatic finite-state machines.

Similarly to pneumatic logic, our work proposes a set of pneumatic flow resistors with a set pressure drop which enables subsequent activation of multi-bladder interconnected chambers through a single pressure input. In contrast to previous flow activation control, our proposed flow-resistors are fully passive without any movable internal parts, and they act as flow restrictive components, this design contributes to the field by introducing a continuous passive element for flow control. Herewith, allowing a set delayed activation based on simple additions in a 3D printed component.

Other works have approached the same problem of decentralizing control for soft actuators in multi-degrees of freedom \cite{Robertson2017,Xu2020} through compact pneumatic regulators for integration within soft-robots \cite{Booth2018}, or through pressure change by orifice plates and multiple geometries studied in different applications \cite{Varoutis2009,Choi2011,Valougeorgis2017}. Similarly, passive valves for flow control in \cite{Napp2014} regulated flow through the use of movable plates assembled for a set pressure, whereas in \cite{Gilbertson2017} single plate orifice valves time-controlled canula robots.

However, no prior work has used a multi-pressure drop in a single passive valve for flow control. The closest principle presented fluid elastomer actuators (FEA) based on porosity in inner layers of the actuators for controlling the motion and timing \cite{Futran2018}.
In contrast, our design proposes resistor valves for continuous interconnected flow, easily fabricated as single components through multiple orifice plate for controlling the fluid flow of any type of inflatable actuators. We introduce its easy implementation and characteristics on series and parallel pneumatic circuits, which results in variable sequential activation based on the circuit configuration without any structural change to the actuators.

\section{Soft Series Inflatable Actuators} \label{s:SIA}
	\begin{figure*}[!t]
	\centering
	\subfigure[Schematic distribution of the embedded SIA on the proposed garment robot-suit.]{\includegraphics[width=6.5cm]{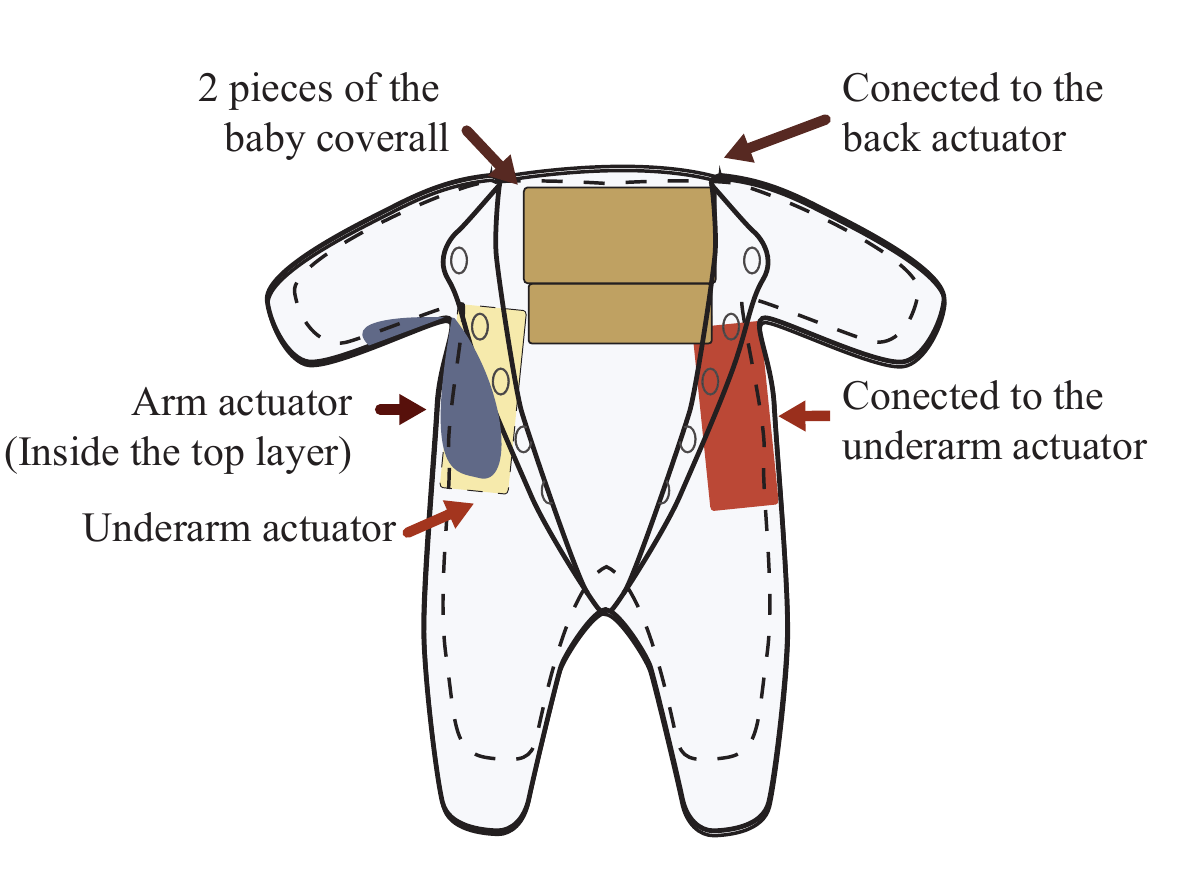}%
	\label{fig:SSIA_schem}} 
	\hfil 
	\subfigure[Experimental setup of the flow control circuit.]{\includegraphics[width=8.0cm]{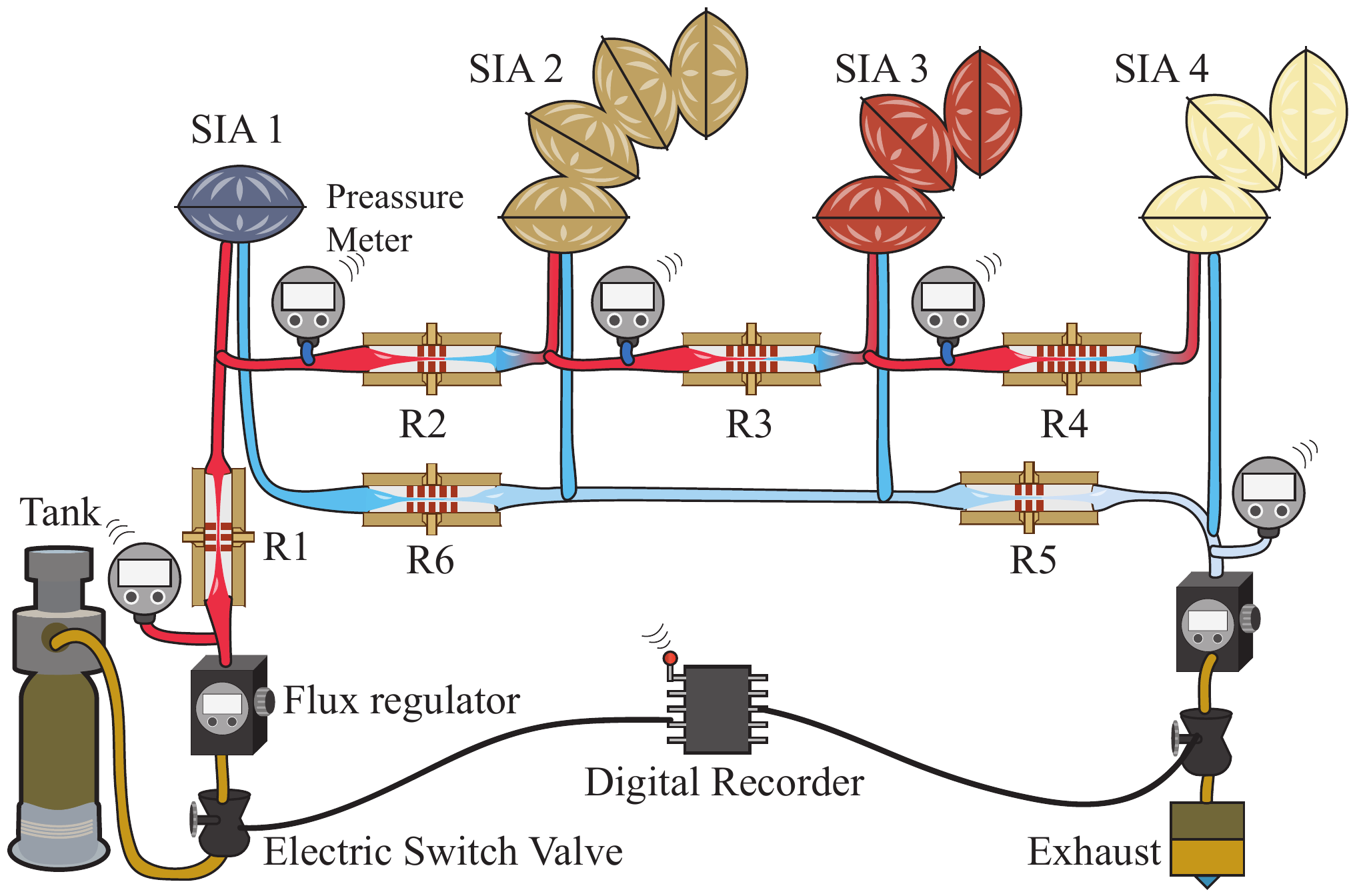}%
	\label{fig:sia_circuit}}
	\caption{Series inflatable actuators with a single inlet valve which initiates the sequential activation through the flow resistors (R1 through R6), the color of each SIA matches the configuration within the robot-suit. 
	\label{fig:SIA_schematic} }
	\end{figure*}
\subsection{Overall Design} \label{ss:design}
The objective application in our case scenario is a soft-robot embedded garment for postural assistance with a set of soft series inflatable actuators, interconnected over multiple limbs of the body, so that they provide assistance to multiple degrees of freedom at a specific sequence of motions. Moreover, we propose the entire system to be interconnected to actuators for controlling the sequential activation from a single pneumatic bypass valve (on/off) for a given input pressure $P_{in}$.

The sequential activation corresponds to the natural postural transition proposed in a symmetric motion with three steps from lying to sitting (LtS), as shown in Fig. \ref{fig:transitions}, proposed in \cite{Yamamoto2015}. Herewith, this method allows a delayed limb motion for long-enough, so that a user of the proposed garment can acknowledge the postural state. 

Initially, for our garment robot we propose a three Degrees of freedom (DoF) configuration with inflatable actuators with sequence in Fig. \ref{fig:transitions}. From an initial lying posture $S0$, the first transition is an arm motion for creating an extended support at the ground in state $S_1$ by SIA-1 with the motion $T_a$. Subsequently, at $S_2$ a rising of the upper body on the sagittal plane is performed with an inclination on the coronal plane executed by SIA-2 at $T_l$ and SIA-3 at $T_f$. Finally, at state $S_3$ elevating the torso on the coronal plane with SIA-4 at $T_r$.

\subsection{Actuators Design}\label{ss:actuators}
\begin{figure}[!t]
      \centering
	    \includegraphics[width=2.7in]{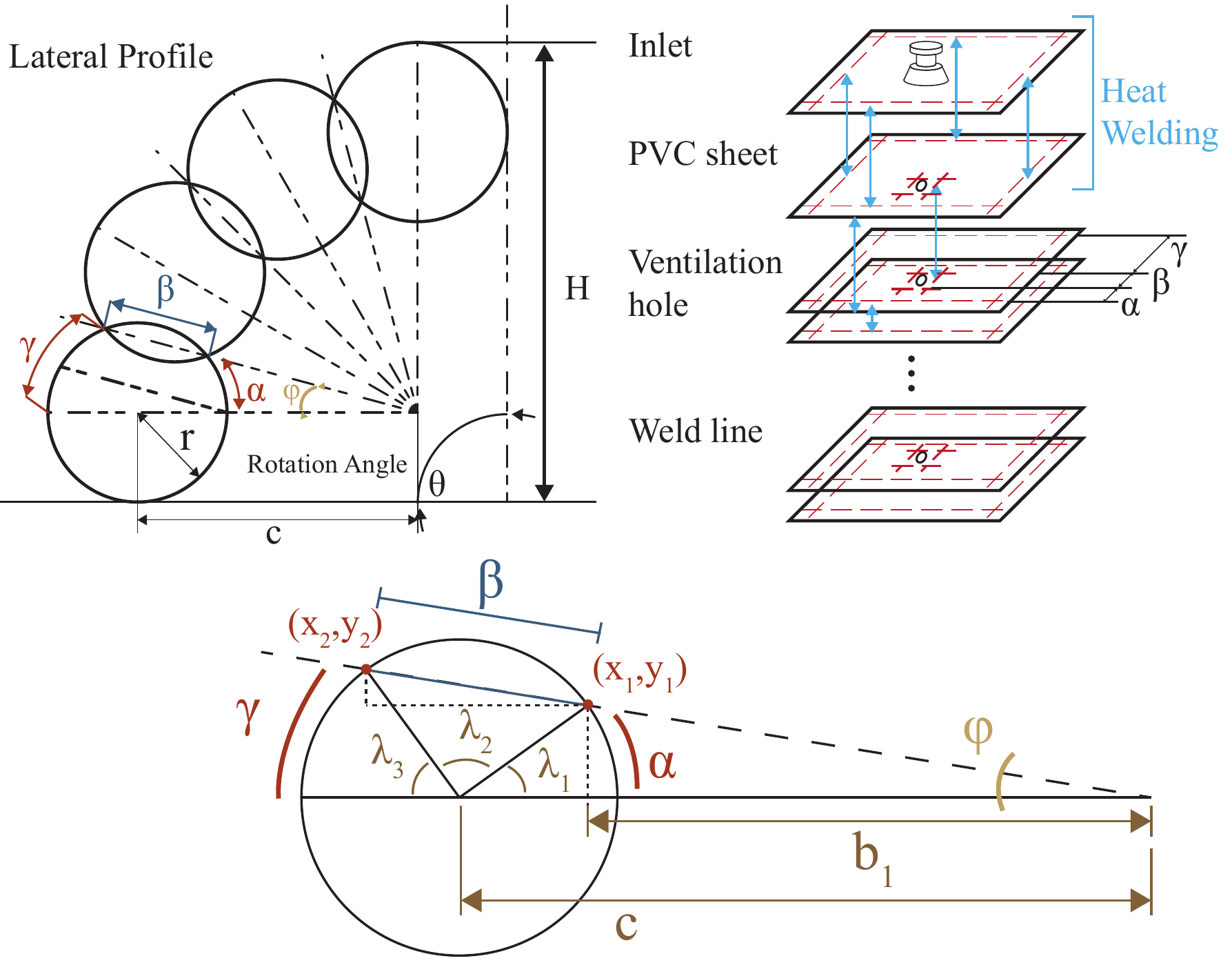}
      \caption{Series inflatable actuators' (SIA) construction for achieving a desired postural transition through embedding on a soft-suit wearable device. \label{fig:SIA_draw}}
   \end{figure}

The garment integrated with SIA, i.e., a set of multi-chambers bladder configured within the soft robot-suit was designed through a combination of multiple layers of semicircles, for achieving a specific range of motion, as depicted in Fig. \ref{fig:SIA_draw}.

We define as design parameter (input) the desired motion range for a given rotary joint $\theta$, achieved at a fully inflated bladder.
Herewith, deriving the construction parameters through the size of the semicircles with a radius $r$, external perimeter portion $\gamma$, internal perimeter $\alpha$, and the interconnection section $\beta$ composing the full actuator by the addition of equally distributed multi-chambers where a single chamber is defined as:
\begin{eqnarray}
    {\phi = \theta/(2n-2) , \label{eq:theta} }\\
     { \lambda_1 + \lambda_2 + \lambda_3 = \pi }, \label{eq:gamma}\\
 	{\alpha = r \lambda_1 , \gamma = r \lambda_3  \label{eq:alpha}}
\end{eqnarray}
where $r$ represents a single in-circle radius, and $\lambda_{1\sim 3}$ denote the inner angles of the intersections on the in-circle, as depicted in Fig. \ref{fig:SIA_draw}. Finally, the perimeters $\alpha$ and $\gamma$ define the main construction parameters.
For a desired output angle $\theta$, height $H$, and actuator radius $r$, we can obtain the construction parameters $\alpha$, $\beta$, $\gamma$, through the intersection of the line that connects the center of rotation $O$ with the circle located at $c$ defined as $(x-c)^2 + y^2 = r^2$ , thus, finding the intersection points $x_{1,2}$ and $y_{1,2}$ for the line $y=x \tan(\phi)$: 
\begin{eqnarray}
    {x_{1,2} = \frac{c \pm \sqrt{\psi}}{1+\tan^2(\phi)}  , \label{eq:x_12} }\\
     {y_{1,2} = \frac{\tan(\phi) (c \pm \sqrt{\psi})}{1+\tan^2(\phi)}  }, \label{eq:y_12}
\end{eqnarray}
where we take $\psi = r^2 + \tan^2(\phi)(r^2+c^2)$ for simpler writing of the subsequent equations. 
Based on these intersection points we can relate the design parameters from the corresponding in-circle angles, thus,  deriving from (\ref{eq:alpha}) the value of $\alpha = \sin^{-1}(y_1 / r)r$, and $\beta$ is derived from the location of the intersection points as:
\begin{eqnarray}
    {\beta^2 = (x_2 - x_1)^2 + (y_2 - y_1)^2, \label{eq:beta1} }\\
    {\beta = \frac{\sqrt{4 \psi (1+\tan^2(\phi)) } }{(1+\tan^2(\phi))^2}  , \label{eq:beta2} }
\end{eqnarray}
from which we defined $\lambda_2$ on the inner isosceles triangle:
\begin{eqnarray}
    {cos(\lambda_2) = 1 - \frac{\beta^2}{2r^2}, \label{eq:lambda2} }
\end{eqnarray}
With this set of equations, a grid search is performed to model all possible solutions for each desired degree of freedom (DOF) and we optimize for a chosen variable, in this case, the number of inner chambers $n$.
   
\subsection{Flow Control Modeling}\label{ss:Modeling}
The proposed soft "flow resistors" were modeled through the gas flow in a medium of multiple capillary-orifices which was found to be comparable to the gas flow through a porous media initially introduced by E.F. Blick in \cite{Blick1966}. This design has been particularly used in hydraulic systems with multiple orifices for flow control when the pressure drop required is too high for single orifices \cite{Tung1983}, but typically no more than 2 or 3 plates were used. To the authors' knowledge, this method of pressure drop control hasn't been used besides pressure sensor devices.
\\
\subsubsection{Resistors Ideal Model}\label{ss:resistors_model}
   \begin{figure}[!t]
      \centering
		\includegraphics[width=2.7in]{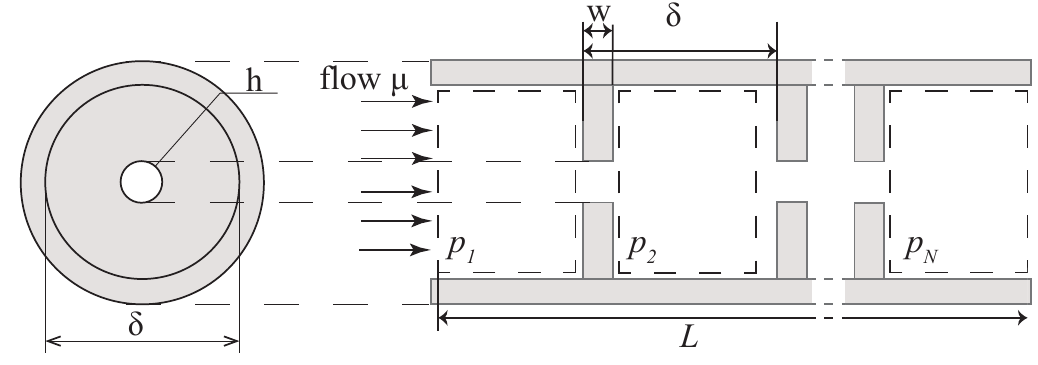}      
      \caption{Design diagram for orifice plate resistor tubes. \label{fig:orifice}}
   \end{figure}
We take the following assumptions to derive the flow within the resistors, from the porous media flow modeled in \cite{Blick1966}:
\begin{itemize}
\item A single fluid is used in the resistor.
\item The orifice diameter is much smaller than the tube mean diameter $\delta>>h$.
\item The thickness of the orifice is much smaller than the mean resistor diameter $w<< \delta$.
\item The flow through the resistor is approximately isothermal, i.e., $T$ at $p_1$ is the same as at $ p_N$.
\end{itemize}

Considering the well known pressure drop for a single orifice plate obtained from the pressure drop at vena contracta ${\Delta p}=p_1-p_h$ the flow can be defined as described in \cite{tadmor2011}:
\begin{eqnarray}
	{ u_i = \xi_i \sqrt[*]{\Delta p}}, \label{eq:flow_1} 
\end{eqnarray}  
where $\xi_i$ denotes a construction semi-empirical constant derived from the area of the orifice plate $A_o$ and tables of drag coefficients $C_D$, with $\xi_i = C_D A_o \sqrt{\frac{2}{\rho}}$, and the operator $\sqrt[*]{x} =  sign(x) \sqrt[]{\vert x \vert}$. For a gas density $\rho$,and a gas flow input velocity  $u_i$.

For multi-capillary flow on a medium like the resistor in Fig. \ref{fig:orifice}, a model was developed in \cite{Blick1966} which found that the total permanent pressure drop was approximately $\Delta p_N = \xi \Delta p$ where $\xi$ was a construction parameter $\xi = (1-\beta^2)$, relating the ratio of the orifice to mean inner diameter $\beta=h / \delta$. Herewith, deriving an orifice coefficient from empirical values of the drag coefficient as:
\begin{equation}
	 	{{C_o}^2 =  {{\left( 1-\beta ^2 \right) \left( \beta ^{-4}-1 \right) } {(2 C_D)^{-1}} } } \label{eq:dP}\\
\end{equation}

This model applies empirical data gathered for fitting a polynomial differential equation $dP/dL = a_1 u_i + a_2 u_i^2$ in terms of the velocity fluid $u_i$ for a given length of the media $L$, and estimates empirical values of $C_o$.

From this experimental equation we take a similar approach, although, instead of $\xi*\Delta p$, we take a single plate orifice $\Delta p_o$ which describes the drop directly by the geometry through $\xi_i$ and include the number of plates $N$ within the resistor by a linear combination,
\begin{equation}
 	{\Delta p_N =  k(N) \Delta p_o}, \label{eq:dP_Sum}
\end{equation}
where we assume $k(N) > 1$ and should be obtained experimentally for a given setup.
The results of our empirical data shows that the best fit for the constrained pressure drop $\Delta p_N = p_{1} - p_{N}$ within a set distance $L$, was $k(N) = \sqrt{N}$, which is detailed in section \ref{ss:physical_resistors}.

\subsubsection{Inflatable Actuator Model:}
In order to estimate the filling $\dot{N}_i$ of an inflatable actuator with volume capacity $V_i$ there are several approaches including finite element models as in \cite{Nguyen2020}, or simpler fitting characteristics of an specific construction through time invariant spring modeling \cite{tadmor2011}. In this work, we take the filling time $T_i = \frac{V_i} {u_i} $ as approximated for a desired operational condition of fluid velocity $u_i$. Therefore, the fitting through a non-linear spring accumulator was performed to find $P_{i} = \sum ^{nf}_{n=1} k_n {(V)}^{n}$, with $n_f$ order polynomial. Assuming a time invariant spring model, meaning no creep or hysteresis affecting significantly.

Subsequently, we can estimate the flux $\dot{N_i}$ through the entire system by the following differential equations:
\begin{eqnarray}
	{ \dot{N}_{1} = f(P_{in},P_{1},\xi_1) - f(P_{in},P_{2},\xi_2) }, \label{eq:flux_1}\\
		{ \dot{N}_{2} = f(P_{1},P_{2},\xi_2) - f(P_{2},P_{3},\xi_3) }, \label{eq:flux_2}
\end{eqnarray}
where the flux through each SIA depends on the pressure of its predecessor $P_{i-1}$ and subsequent $P_{i+1}$ serially connected actuators. In an ideal flow control for any number of actuators the flux is defined as:
\begin{equation}
	{ \dot{N}_{i} = f(P_{i-1},P_{i},\xi_{i}) - f(P_{i},P_{i+1},\xi_{i+1}) }, \label{eq:flux_i}\\
\end{equation}
where the filling velocity is defined by the construction parameter of the orifice plate $\xi_n$ and the number of plates $n$ within the resistor. These parameters can be used directly for a forward design through optimization for a desired time delay given a known actuator volume $V_i$, the actuator equation and the expected operational pressure $P_{in}$.

\section{Physical Realization and Characterization}\label{ss:physical_resistors}

\subsection{Flow resistors construction and characterization:} \label{ss:resistors_construct}
We design the flow resistors within a single tube of length $L$, with inner diameter $\delta$, composed of $N$ number of orifice plates of thickness $w$, and orifice diameter $d$ as shown in Fig. \ref{fig:orifice}.
The procedure for design follows the top-bottom process:
\begin{enumerate}
	\item Select connector diameter $\delta$.
	\item Select orifice plate diameter $h$ and thickness $w$ so that it satisfies the conditions set in section \ref{ss:resistors_model}.
	\item Select $L$, so that $N$ plates fit within $L$
\end{enumerate}
We envisioned the flow resistors as simple passive components for usage with standard air pressure valves and tubes. Therefore, we chose characteristics for fitting standard tubing through one-touch pipes. i.e., a desired external diameter $D=6mm$ in the resistors and length of $L=40mm$. 
Subsequently, the size and distribution of the orifice followed the guidelines set in section \ref{ss:Modeling}, with $\delta = 4 [mm]$, an $h = 0.98 [mm]$, and $w = 0.5 [mm]$, giving a $\beta=0.245$, and building up to $N = 7$ orifice plates considering a minimum spacing larger than $\delta$.
\begin{figure}[!t]
      \centering
		\subfigure[Flow resistor's CAD design: example of a 3 orifices resistor.]{\includegraphics[width=8.0cm]{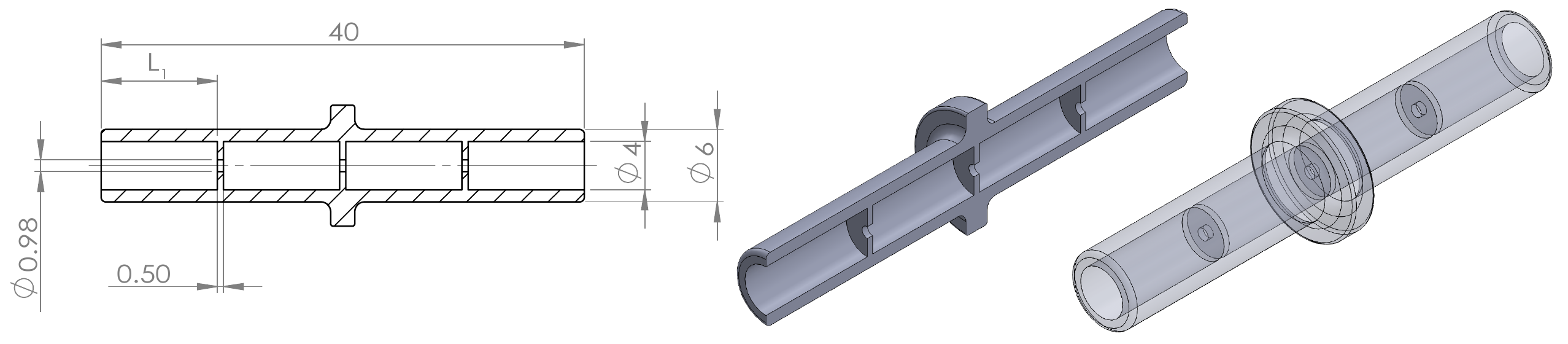}%
		\label{fig:tubeCAD}}
			\hfil 
		\subfigure[Air flow resistors: Left to right one to seven orifice plates.]{\includegraphics[width=6.0cm]{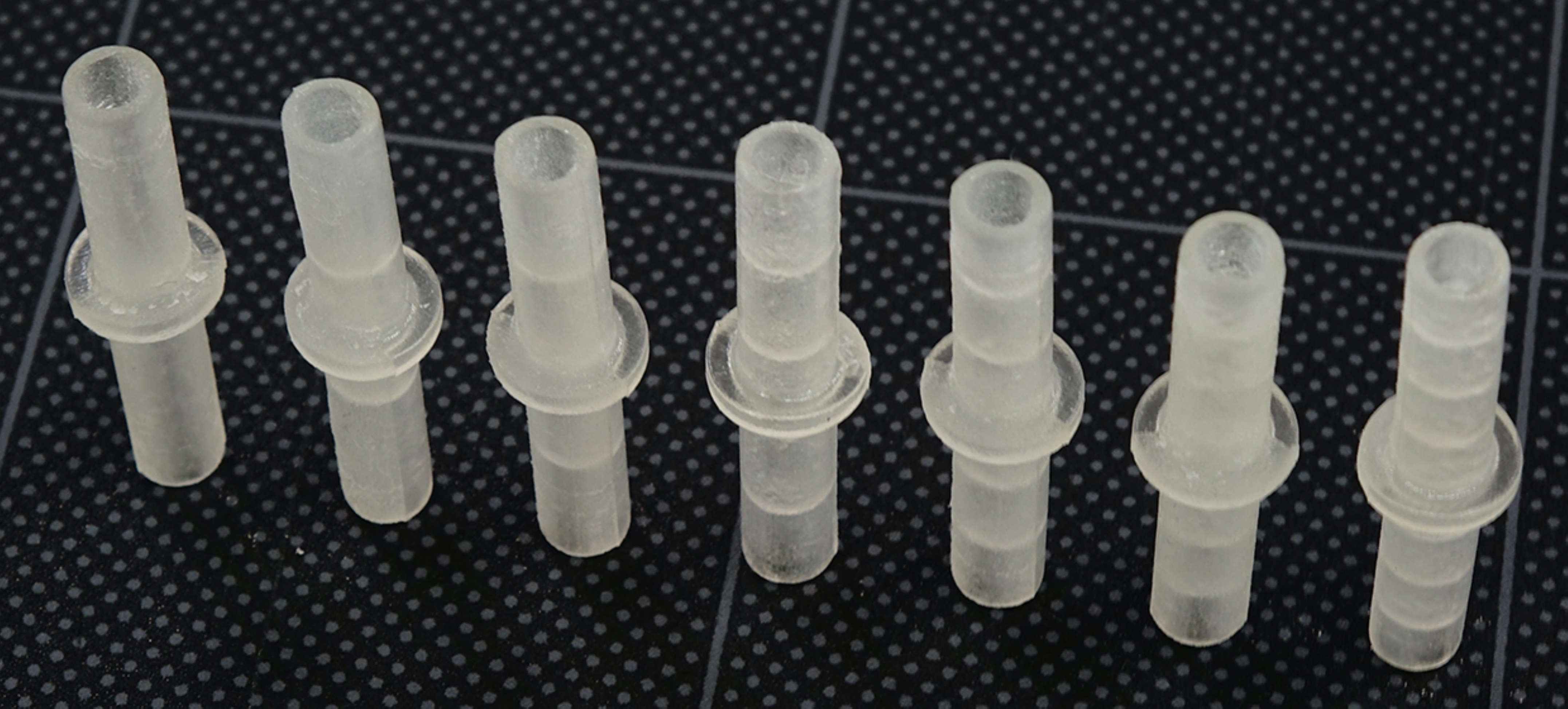}
		\label{fig:tubes_pic}}	
      \caption{Series Inflatable actuators construction and flow resistor configuration for achieving the desired postural transitions.
      \label{fig:tubes} }
   \end{figure}
   
The material for the fabrication was selected for structural strength and flexibility in a 3D printer Objet350 Connex 2 (from Stratasys, USA) a combination of tangoPlus (FLX930) and VeroClear (RGD810) materials producing a digital material FLX9985-DM with elongation break rated at $60\%$, tensile strength of $7.0 Mpa$, tensile tear resistance of $25 Kg/cm$, and shore hardness of 85A. The latter being comparable to industrial standards of polyurethane tubing for pressures under 135 PSI, typically A82 to A98 (ISO/TS 11619:2014). Fig. \ref{fig:tubes} shows the printed resistors and the CAD model.

\begin{figure}[!t]
      \centering
		\subfigure[Measured activation pressure of the inner actuation chambers with different flow resistors, highlighting the $\Delta_{p}$ differences to the input pressure $P_{in}$, .]{\includegraphics[width=8.0cm]{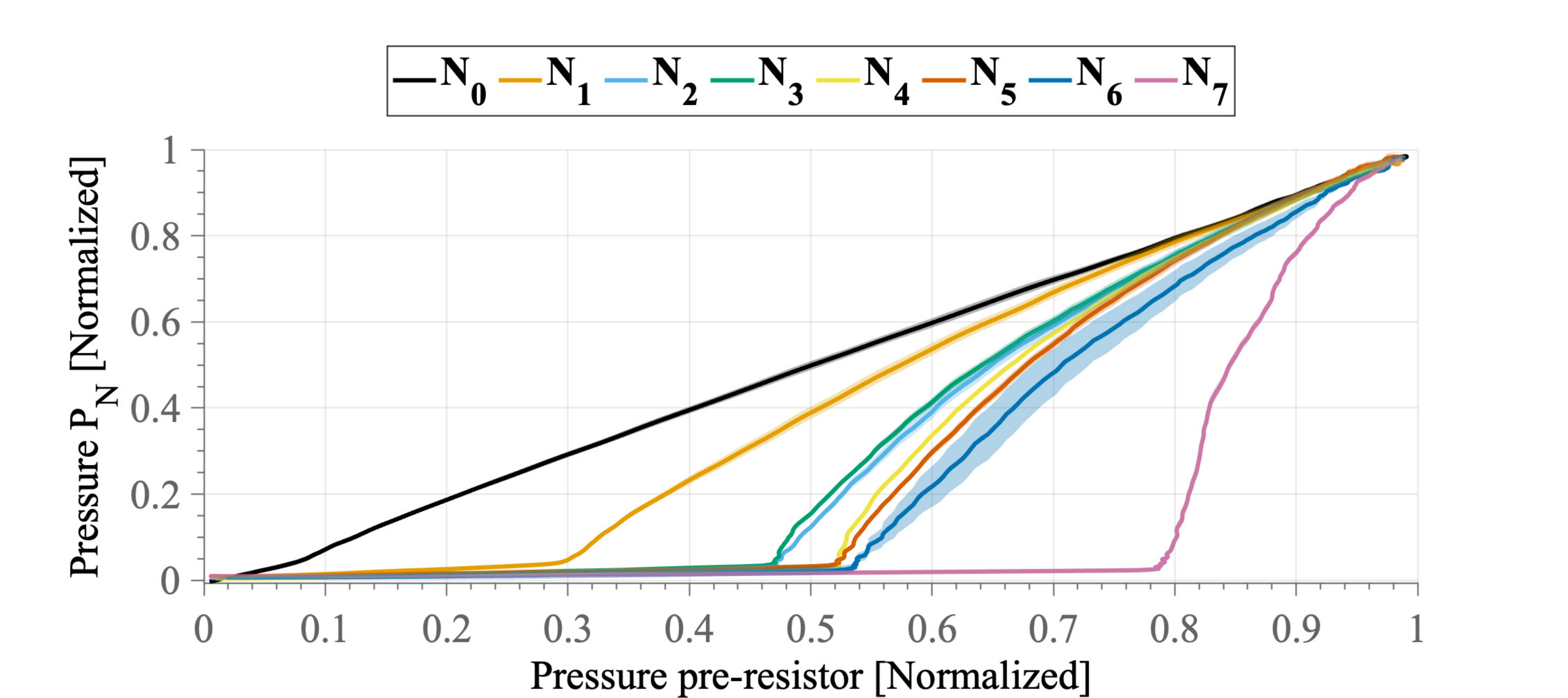}%
		\label{fig:measure_drop}}
		\hfil 
		\subfigure[Measured pressure change compared to pre-resistor pressure.]{\includegraphics[width=8.0cm]{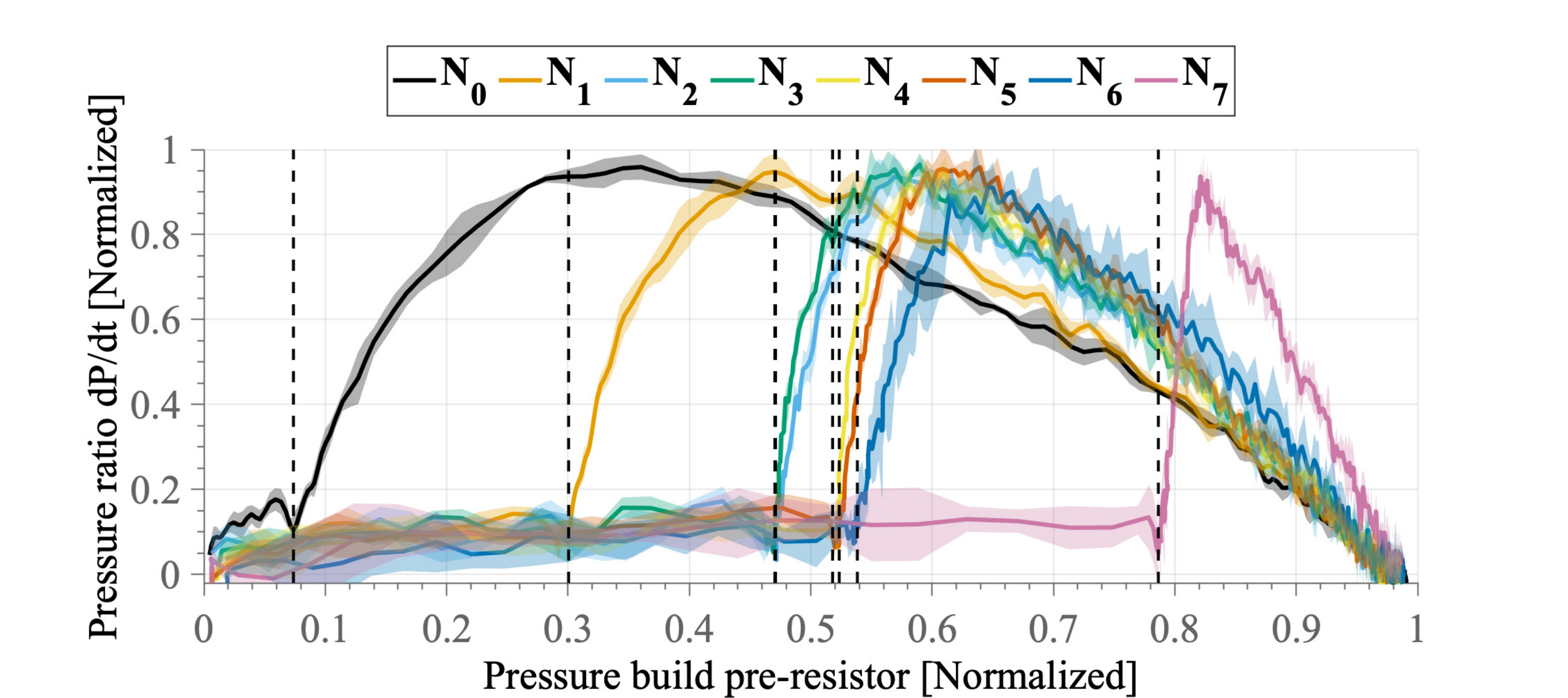}%
		\label{fig:pressure_change}}
		\hfil 
		\subfigure[Measured time delay when activating individual chambers with different flow resistors.]{\includegraphics[width=8.0cm]{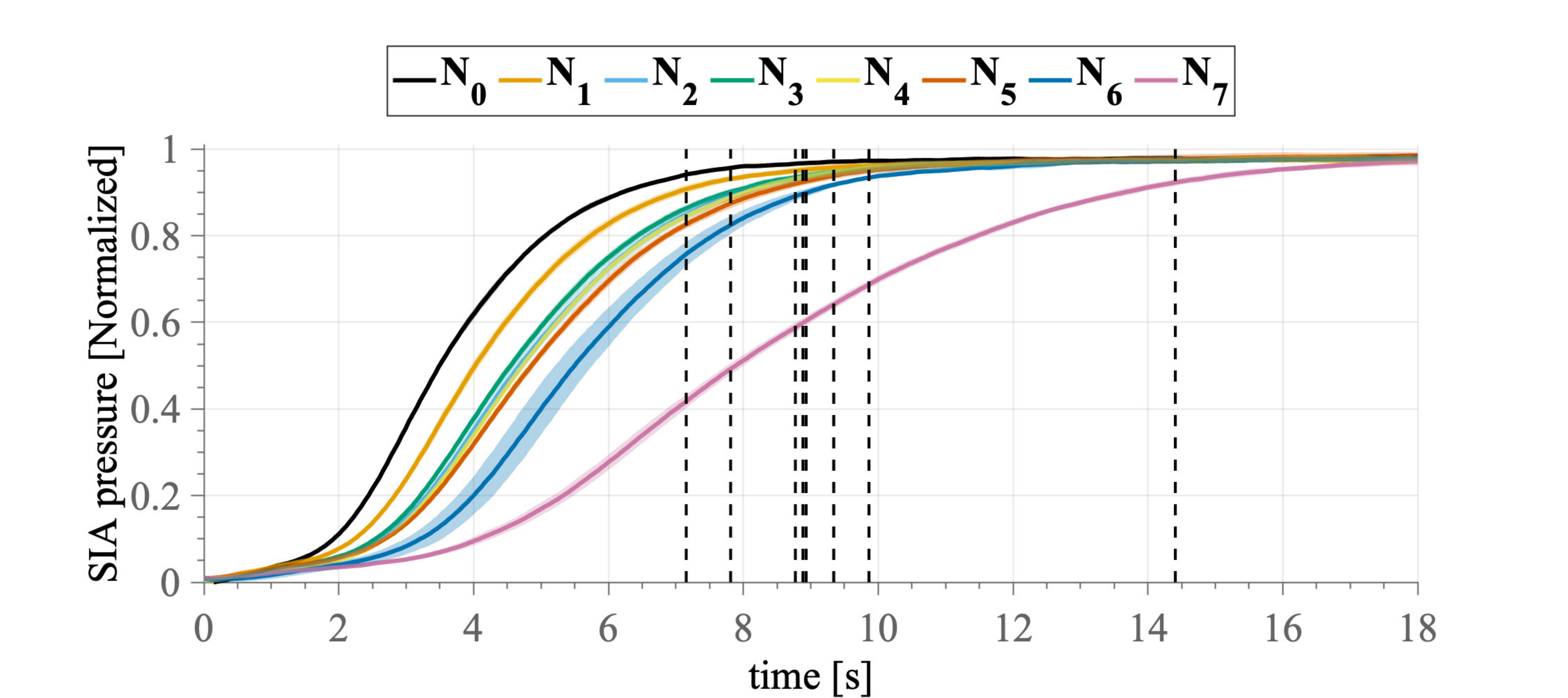}
		
		\label{fig:time_delay}}
      \caption{Testing of series inflatable actuators with individual orifice plates resistors. The results show the drop in pressure for each resistor in four round of tests for each configuration.
      \label{fig:resistors_individual} }
   \end{figure}

We evaluated individual flow resistors for inflating and deflating stages by automatically controlling the setup with input and escape bypass solenoid valves while recording pressure changes through a set of sensors MIS-2503-150G (Metrodyne Microsystem Corp. Taiwan). 
The assessment and characterisation was performed focusing on the inflating control with a pressure input of $P_{in}=50 kPa$ controlled by a flow valve, recorded 4 times per single actuator. The results are shown in Fig. \ref{fig:resistors_individual}, with all measurements at the actuator's chamber $p_N$ normalized to the maximum $P_{in}$.

We compared the pressure drop $\Delta p_N$ for increasing number of orifice plates ($N$) in the real implementation (Fig. \ref{fig:measure_drop}) with the assumptions made in the formulation in (\ref{eq:dP_Sum}). Here, we studied the pre-resistor pressure $p_1$ against the actuator pressure $p_N$ (shown in Fig. \ref{fig:orifice}), and used the time derivative on each measurement $dP/dt$ for consistently taking $\Delta p_N$ at the inflection of activation pressure (shown in Fig. \ref{fig:pressure_change}) i.e., the moment where the flux $\dot{N}_i$ was comparative for all tests. 

\begin{figure}[!t]
      \centering
		\includegraphics[width=7.0cm]{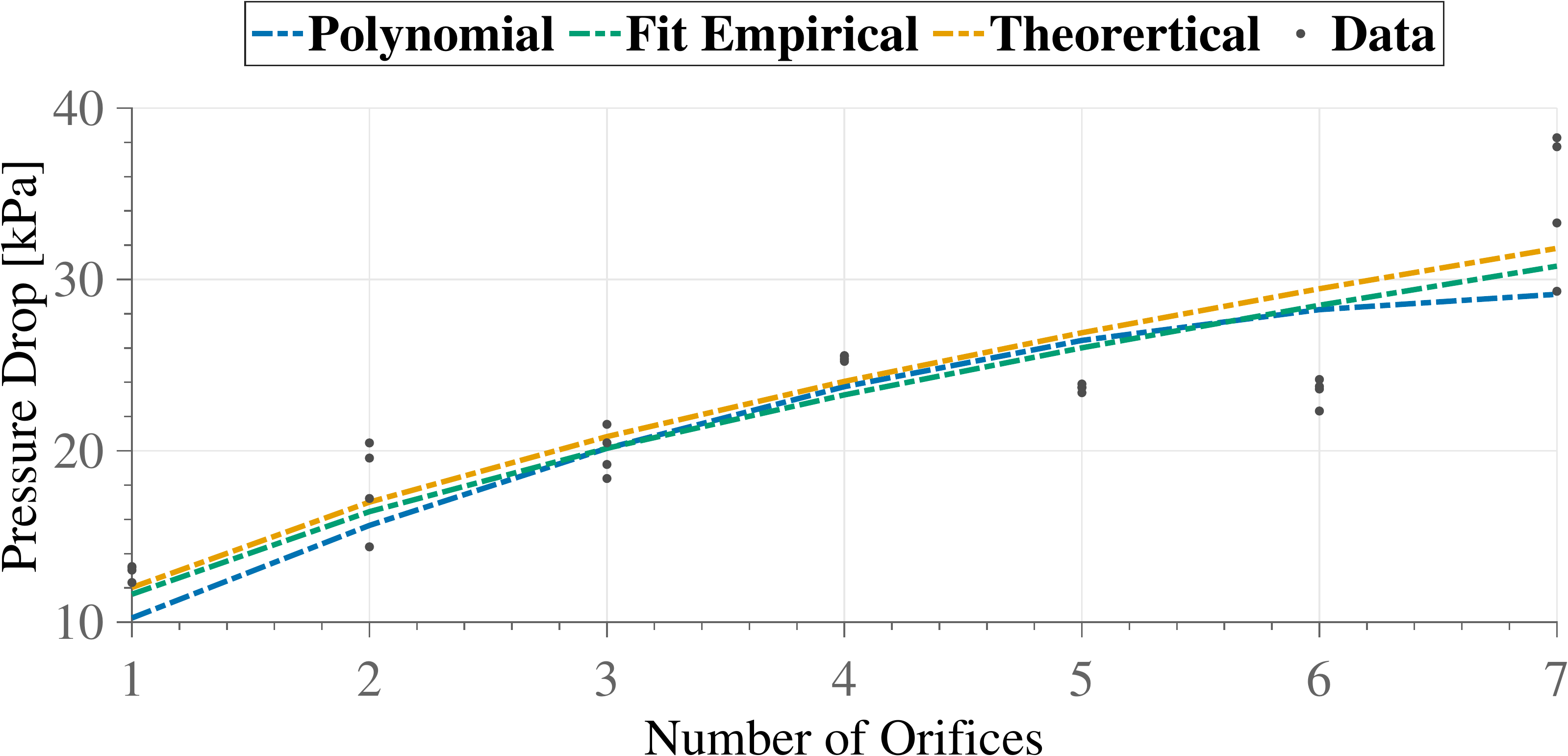}      
      \caption{Theoretical model of the pressure drop given by (\ref{eq:dP_Sum}) compared with a linear fit of the experimental evaluation of the flow resistors. \label{fig:pressure_drop}}
   \end{figure}

We summarise in Fig. \ref{fig:pressure_drop} different fitting options for the experimental value of the pressure drop. Comparing a second order polynomial fit $k(N) = \sum ^{2}_{n=1} a_n {(N)}^{n}$ which showed a $RMSE=3.94$ with $R
^2=0.84$, whereas an equation of the form $k(N) = a*\sqrt{N}$ yielded a $RMSE=3.1$ with $R
^2=0.89$ making it a better fit. Moreover, the theoretical dependence only in the geometry of the resistor data with $k(N) = \sqrt{N}$ shows comparable fitting results $RMSE=3.15$ with $R^2=0.88$, which represents an error of $6.1 \%$ for the measurements. 
Although the data showed some variability in the activation pressure with the number of plates in this assessment, as seen for $N_5$ and $N_6$, we observed in the interaction with the actuator that the filling time did increase with the number of plates, which we describe in table \ref{table:times} and detailed in the next section.
   
\subsection{Inflatable Actuators Activation:}
Here, we assessed the time to reach full inflation on a single inflatable actuator in combination with the flow resistors. We set inflation measure at $95\% P_{in}$, with mean, standard deviation (STD) and time difference to a baseline no-resistor $\Delta t_N = t_N - t_0$, presented in table \ref{table:times}.

\begin{table}             
    \centering                
    \begin{tabular}{c|c|c|c|c}    
    \hline                    
    \hline
    Flow Resistor & Mean [s] & STD [s] & $\Delta t$ [s] & $\Delta P_{N}$ [kPa] \\
    \hline
    \hline
    N0 & 7.15 & 0.11 & 0.00 & 2.19 \\
    \hline
    N1 & 7.81 * & 0.05 & 0.66 & 12.95 \\
    \hline
    N2 & 8.93 * & 0.09 & 1.78** & 17.91 \\
    \hline
    N3 & 8.77 & 0.04 & 1.62** & 19.90 \\
    \hline
    N4 & 8.88 & 0.12 & 1.73 & 25.41 \\
    \hline
    N5 & 9.34 * & 0.09 & 2.19** & 23.67 \\
    \hline
    N6 & 9.86 & 0.42 & 2.71 & 23.46 \\
    \hline
    N7 & 14.41 * & 0.12 & 7.25** & 34.66 \\
    \hline
    \hline                    
    \end{tabular} 
\vspace{1ex}
    {\raggedright \\ $*$ marks significant difference at $p<0.01$ of $N_i - N_{i-1}$ and $**$ marks the same level of significant difference for $N_i - N_{i-2}$ . \par}
\caption{Time table for reaching $95\%$ of the input pressure, and pressure drop at the initial inflection point.}
\label{table:times}
\end{table}   

We evaluated the potential application for timed control through a one-way ANOVA for each resistor couple $N_i-N_{i-1}$, with results showing significant time difference at the level $p<0.01$ marked with $*$ on table \ref{table:times}. As well, we observed the significant difference at the level at $p<0.01$ for the couples $N_i-N_{i-2}$, marked as $**$. Although all resistors perform as expected in delaying activation of the SIA, the significant difference among themselves was found only for some resistors in our current setup $N_1$, $N_2$, $N_3$, $N_5$, $N_7$. Therefore, we selected and evaluate some of them in the following section.

\subsubsection{Characterization of circuit connections:}
\begin{figure}[!t]
      \centering
      	\subfigure[Schematic circuits comparing the sequential activation of series and parallel configurations.]{\includegraphics[width=8.5cm]{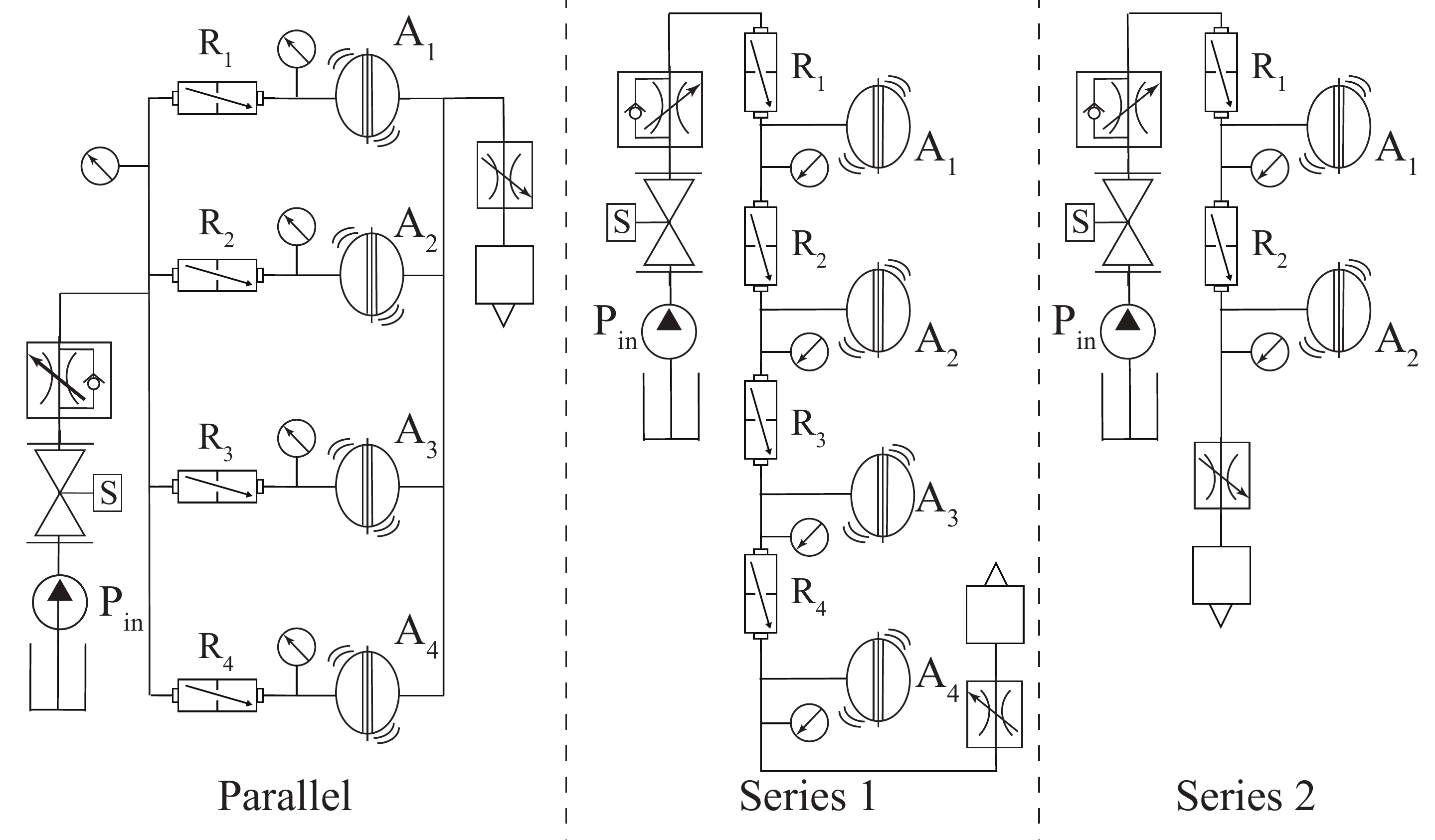}%
		\label{fig:circuits_schematic}}
		\hfil 
        \subfigure[Experimental setup.]{\includegraphics[width=4.2cm]{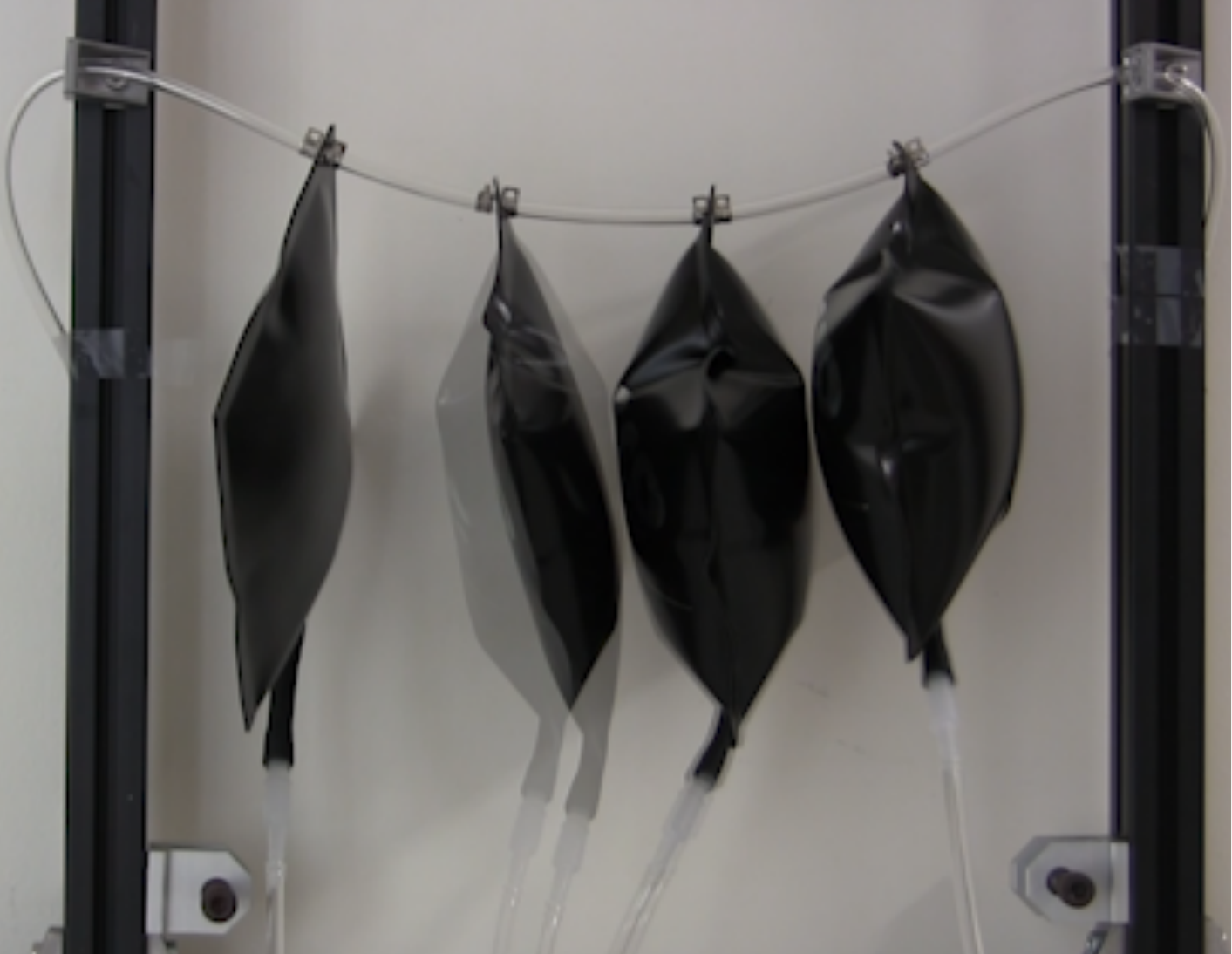}%
		\label{fig:circuits_photo}}
		\hfil
		\subfigure[Series circuit.]{\includegraphics[width=4.2cm]{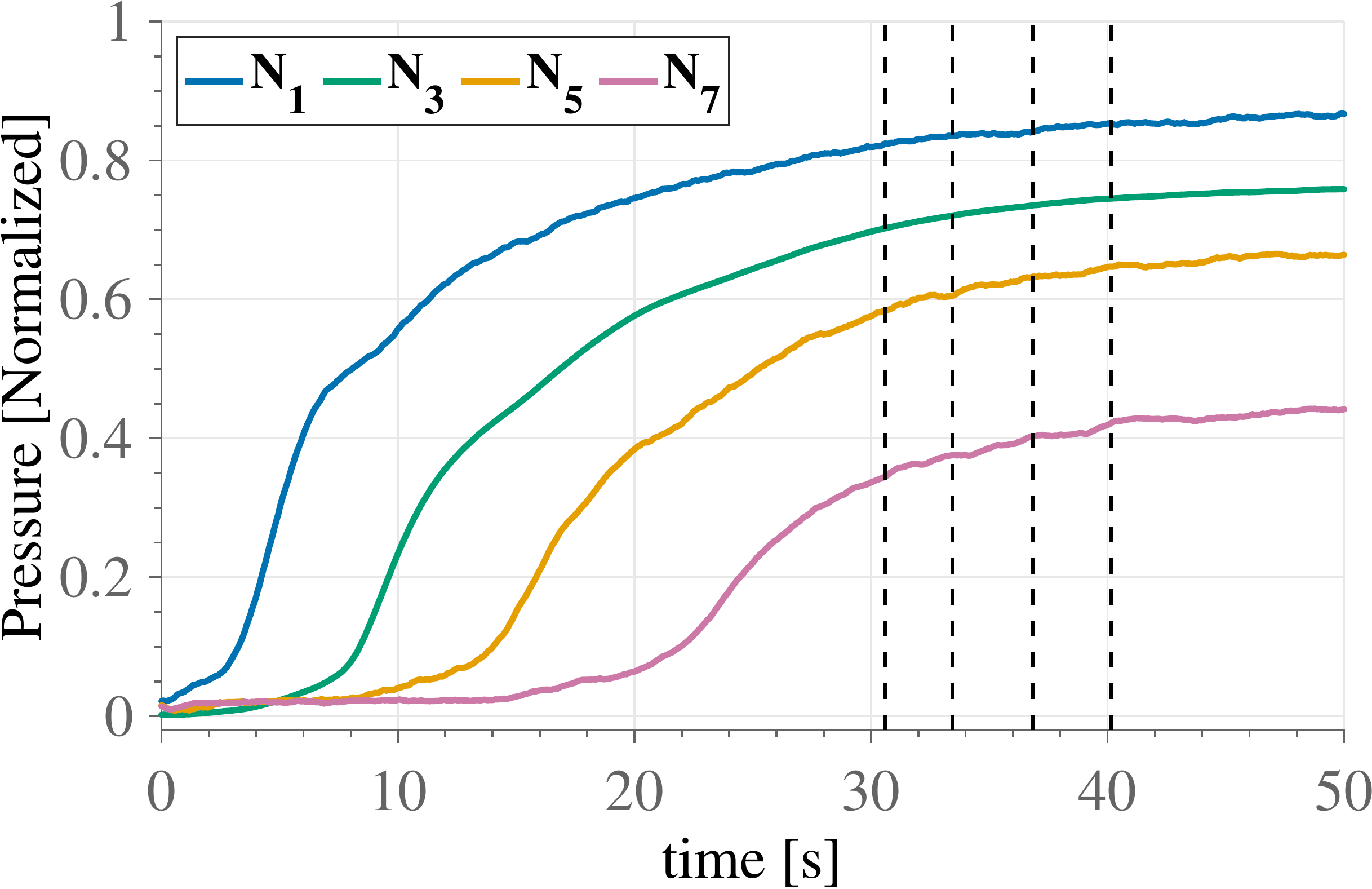}%
		\label{fig:res_series_cont}}
		\hfil
		\subfigure[Parallel circuit with closed flow.]{\includegraphics[width=4.2cm]{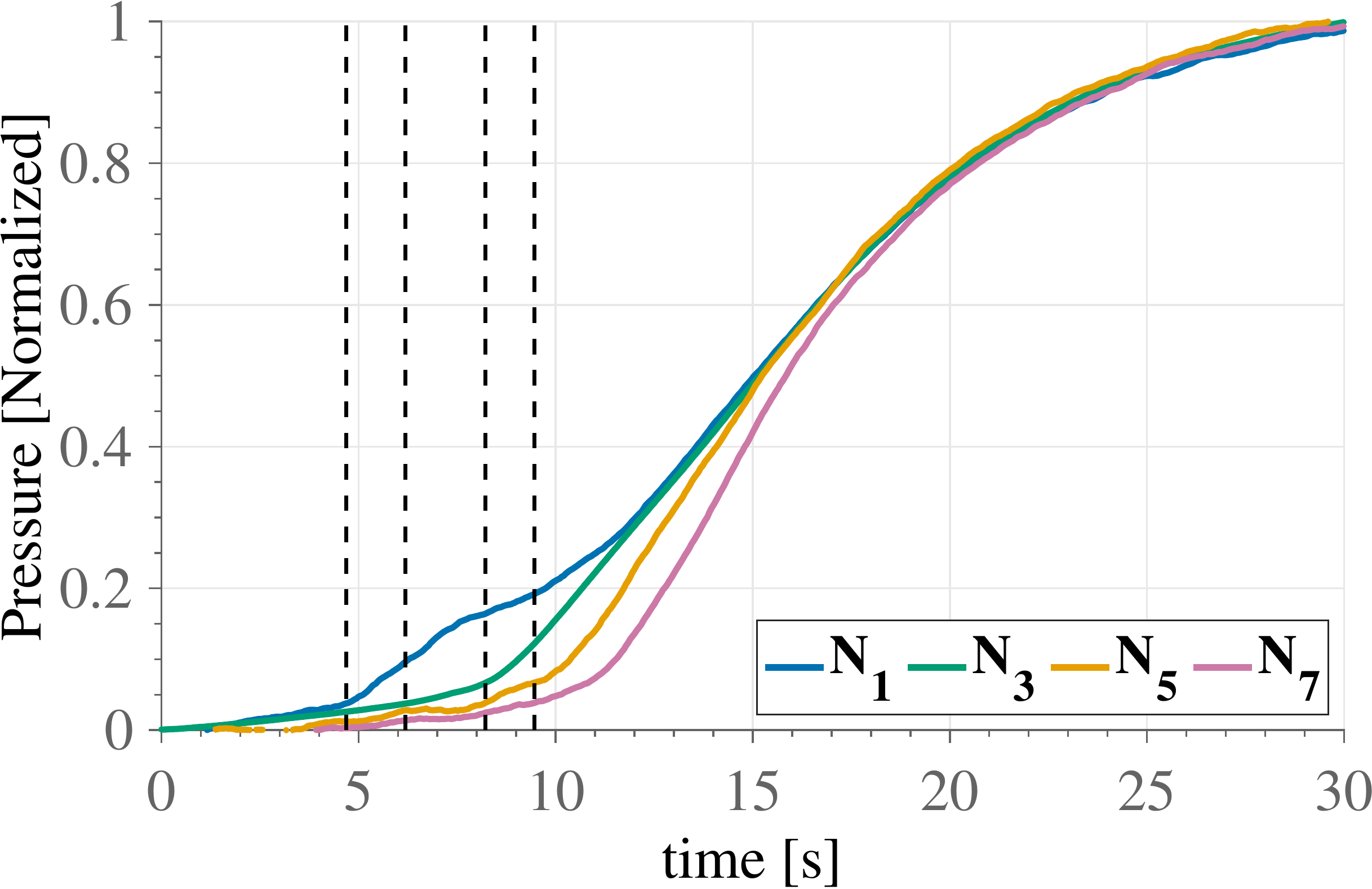}%
		\label{fig:res_parallel_block}}
		\hfil
        \subfigure[Series circuit with closed flow.]{\includegraphics[width=4.2cm]{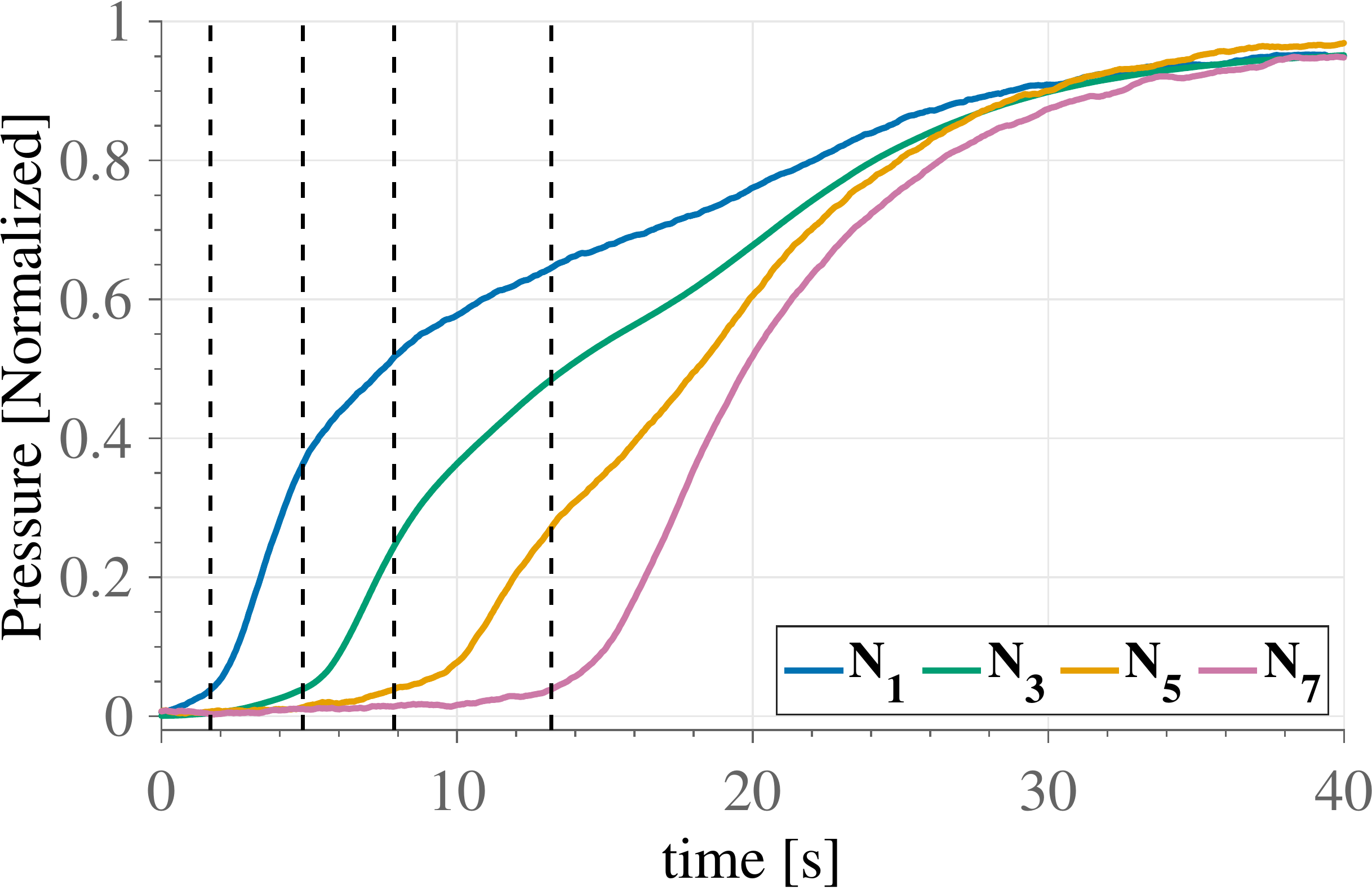}%
		\label{fig:res_series_block}}
		\subfigure[Series 2 circuit, testing $R_1=N_3$ followed by $R_2=N_5$.]{\includegraphics[width=4.2cm]{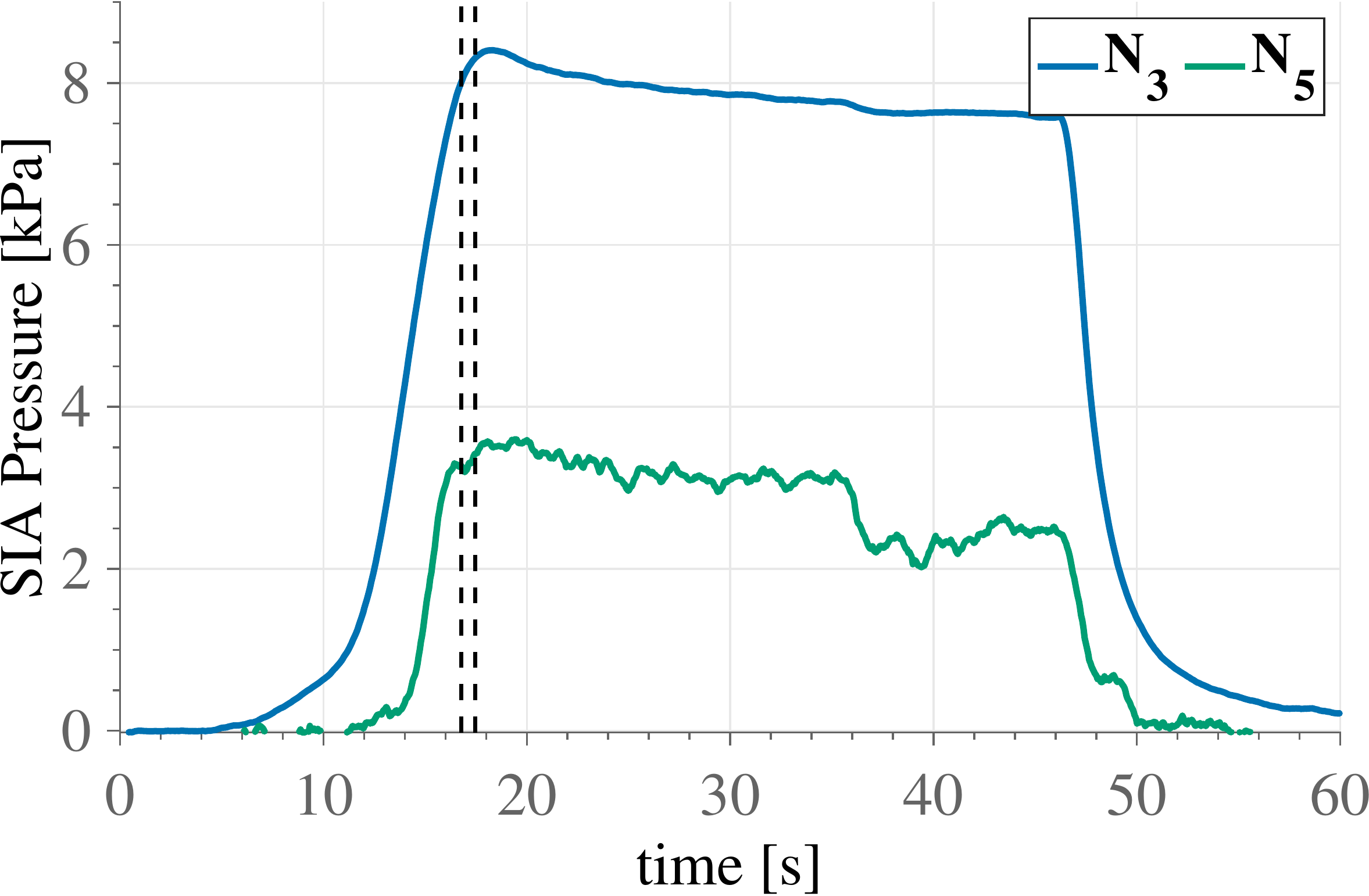}%
		\label{fig:activation_3-5}}
		\hfil 
		\subfigure[Series 2 circuit, testing $R_1=N_5$ followed by $R_2=N_3$.]{\includegraphics[width=4.2cm]{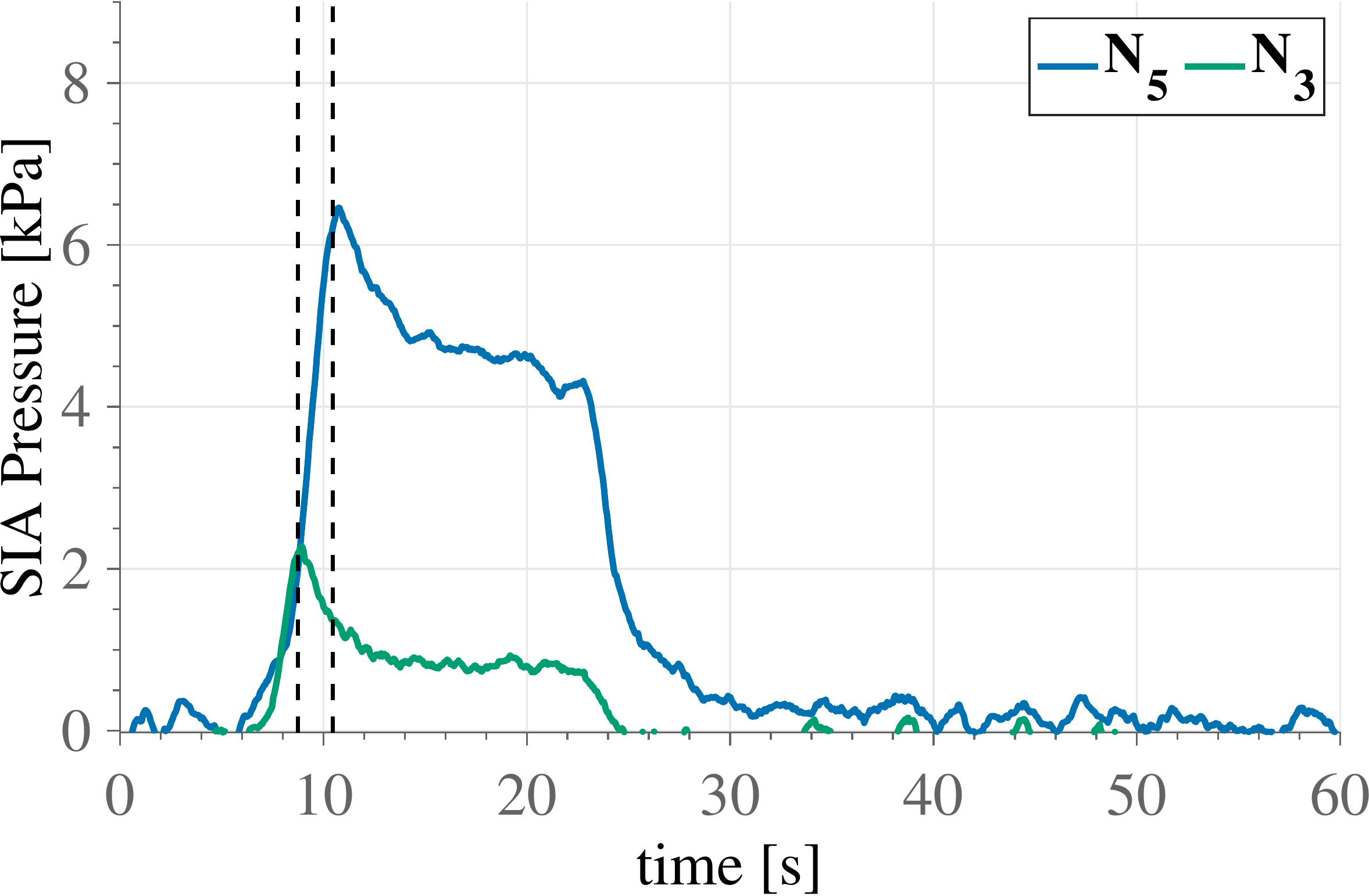}%
		\label{fig:activation_5-3}}
      \caption{Series and parallel circuits of the flow resistors, highlighting the pressure over time in equal inflatable actuators when connected to a single input pressure and single on/off bypass valve (please refer to the attached multimedia file for these recordings).
      \label{fig:test_circuits} }
   \end{figure}

We evaluated the implementation on two circuit configurations: series and parallel as shown in Fig. \ref{fig:circuits_schematic}.

Fig. \ref{fig:res_series_cont} shows continuous air flow with a single inlet pressure valve at $50 kPa$ and a continuous flow through the outlet set at $20kPa$. The pressure drop on each SIA corresponds to $[0.13, 0.11, 0.09, 0.22]$, which corresponds to an activation lag for $N_3$ of $2.84s$, $N_3$ of $3.39s$, and $N_7$ of $3.28s$. These behavior is as expected for a subsequent set of SIA interconnected with subsequent pressure drop given for each resistor which is a value proportional to the differential pressure in $P_i$ and $P_{i-1}$.
For comparison we present the same configuration with a closed outlet in Fig. \ref{fig:res_series_block}, and a parallel circuit in Fig. \ref{fig:res_series_block}. In this case, we observe still a sequential activation of the actuators in the ascending order $N_1$ through $N_7$ for series circuit a $\Delta t$ at activation $3.0s$, $3.1s$, and $5.2s$ for $N_3, N_5, N_7$, whereas in the parallel circuit a much smaller difference is given to $\Delta t$ at activation $1.4s$, $2.0s$, and $1.24s$ for $N_3, N_5, N_7$. In both cases reaching the full inflation at virtually the same time.

Subsequently, we characterized the response in a series circuit of two SIAs (Fig. \ref{fig:circuits_schematic}, right side) with $R_1=N_3$ followed by an $R_2=N_5$ (Fig. \ref{fig:activation_3-5}) compared with the inverted resistor's order in Fig. \ref{fig:activation_5-3}. In the first setup $N_3$ followed by $N_5$ the sequential activation shows a $\Delta t=0.67s$ with a consecutive activation of the SIA as all previous circuits. On the other hand, the inverted sequence $N_5$ followed by $N_3$ shows a $\Delta t = -1.7s$ i.e., SIA-2 reached its maximum pressure at a lower value before the predecessor SIA-1. This response resulted from the pressure built on SIA-1 reaching the subsequent resistor's ($R_2$) activation pressure faster because of the lower required pressure ($\Delta P_3$) as presented in table \ref{table:times}.

The result of this characterization means that during the design process, the desired sequence and activation pressures of the SIA must be set in order to construct a circuit that fulfils both criteria. A forward dynamic simulation following the differential equation in (\ref{eq:flux_i}) could be used to estimate the activation time and through (\ref{eq:dP_Sum}) estimate the SIA pressures. e.g., on a support system over the arm, one could set an activation sequence of an elbow-joint followed by a shoulder-joint, with the elbow flexion at a lower pressure (smaller joint torque required) followed by a shoulder abduction at a higher pressure (which should require a higher torque) all through a circuit like $N_5 - N_3$ in Fig. \ref{fig:activation_5-3}.

\section{System Evaluation}\label{sec:eval}

\subsection{Soft Wearable Suit Construction:} \label{ss:construction}
The current implementation of the wearable garment was built for fitting a dummy resembling a 1-year-old child, accordingly to the body parts distribution in \cite{Mochi2013} with a total weight of $7.95 [kg]$ and $0.82 [m]$.

\begin{figure}[!t]
      \centering
  		\subfigure[Construction sample of a 4-layered SIA.]{\includegraphics[width=7.2cm]{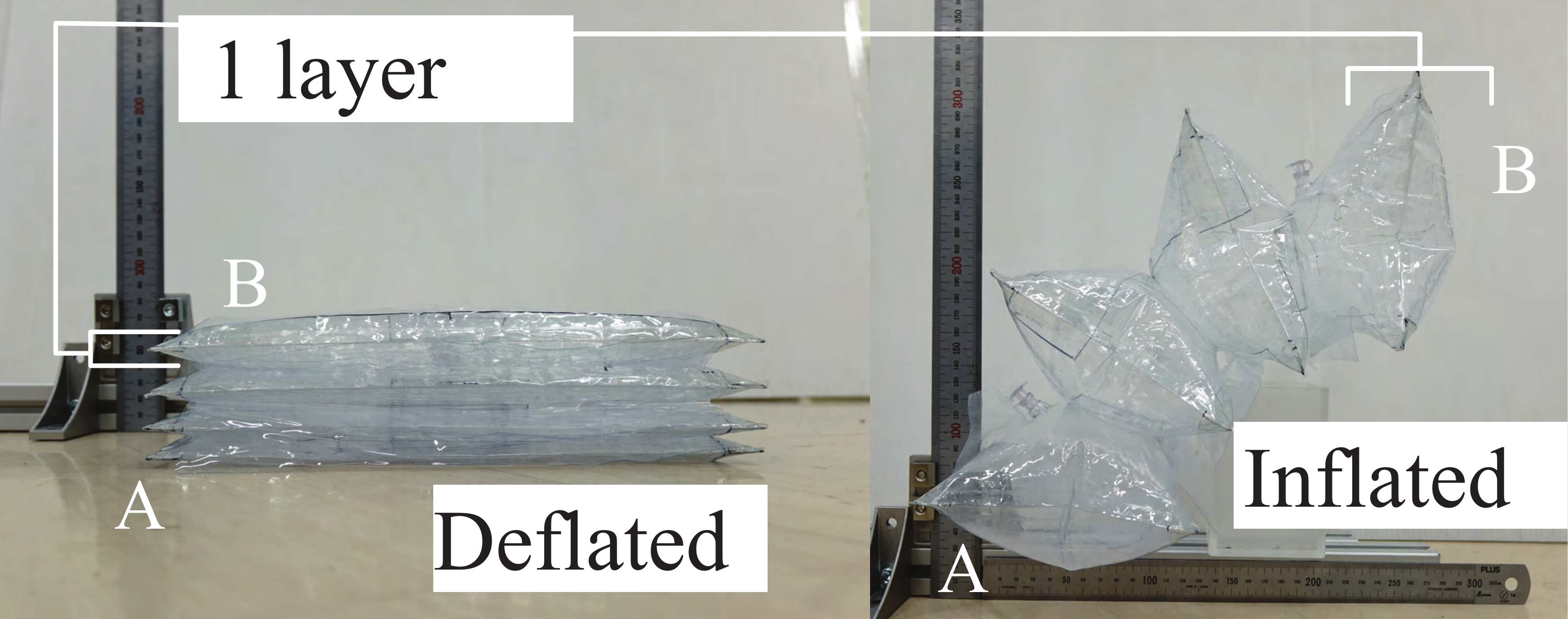}
		\label{fig:SIA_const}}
		\hfil
		\subfigure[SIA distribution over the soft-suit for postural assistance.]{\includegraphics[width=8.6cm]{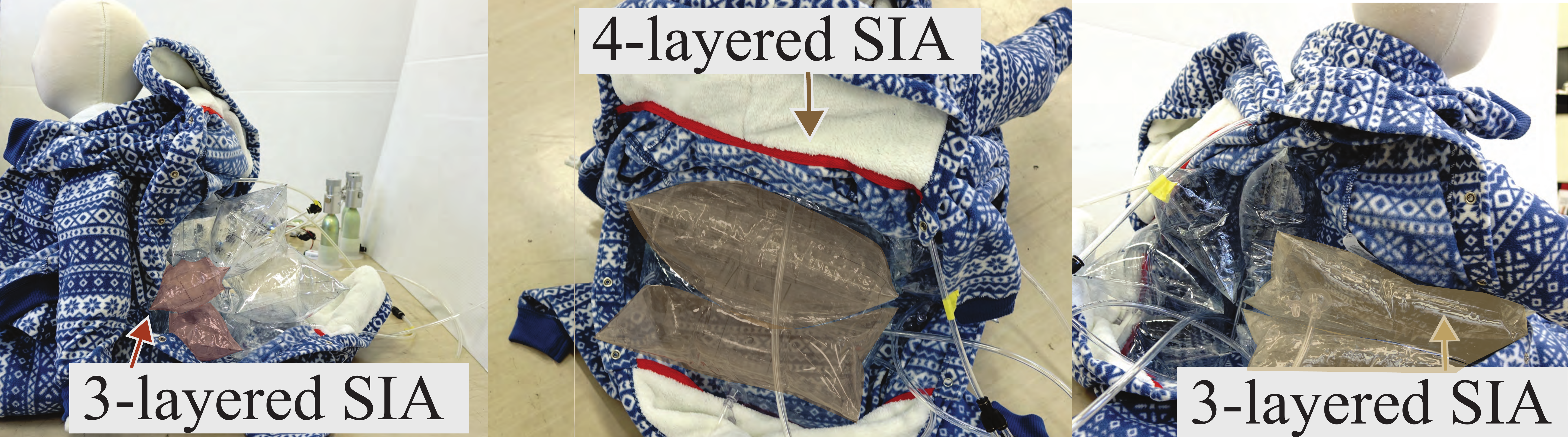}%
		\label{fig:SIA_suit}}
      \caption{Series Inflatable actuators construction and flow resistor configuration for achieving the desired postural transitions.
      \label{fig:SIA} }
   \end{figure}
The actuators were built using polymerizing vinyl chloride (PVC) film of $0.1 [mm]$ following the schematic design in \ref{fig:SIA_schematic}, a one layer inflatable actuator was set at the right arm for the first transition (SIA-1), a 4 layered (SIA-2) for supporting the back, and a 3 layered SIA for each lateral support of the torso (SIA-3 and SIA-4), as shown in Fig. \ref{fig:SIA_suit}.

We used an exhaustive exploration in Matlab (Mathworks Inc. USA) for the desired height $H$ and rotation angle $\theta$ with a given chamber size $r$ for obtaining the parameters $\alpha$, $\beta$, and $\gamma$ to fit a number of layers $n$ with the formulation given in section \ref{ss:resistors_construct}, while the width of the actuator $W$ was chosen larger than the height. The results are given in table \ref{tab:SIA}.

\begin{table}[t]
    \centering                
    \begin{tabular}{c|c|c|c|c|c|c|c}    
    \hline                    
    \hline
    Actuator & Layers & $\theta [^{\circ}]$ & $H$ & $W$ & $\beta$ & $\alpha$ & $\gamma$ \\
    \hline
    SIA 1 & 1 & $90$ & $60$ & $350$ & - & - & - \\
    \hline
    SIA 2 & 4 & $90$ & $250$ & $300$ & $51$ & $12$ & $58$  \\
    \hline
    SIA 3 & 3 & $90$ & $150$ & $300$ & $61$ & $12$ & $44$ \\
    \hline
    SIA 4 & 3 & $90$ & $150$ & $200$ & $61$ & $12$ & $44$ \\
    \hline
    \hline                    
    \end{tabular} 
    \vspace{1ex}
    {\raggedright \\ All Units in [mm] except $\theta$. \par}
    \caption{Series inflatable actuators construction setup for the wearable transition support suit}
    \label{tab:SIA}
\end{table}   

\subsection{Testing on Postural Transition:} \label{sec:application}

The sequential activation of SIA-1 through SIA-4 was constructed with a series circuit as shown in Fig. \ref{fig:SIA_schematic} from $R_1$ to $R_4$ control the inflating of the suit and set the motion for LtS in the body (baby dummy) with activation of SIA-1 for the right shoulder motion, SIA-2 for trunk rise, SIA-3 for right-side trunk rotation, and finally SIA-4 for trunk rotation to the left side. Resistors $R_5$ and $R_6$ control the descent in the StL transition by first deflating SIA 4 (right side of the suit), followed by the back actuators (SIA 2 and 3), and finally releasing pressure at the underarm actuator SIA 1.
In this testing the input pressure was set to $P_{in} = 100kPa$ and regulated exhaust pressure at $P{out} = 7 kPa$.

Assessment of the motion was performed by a motion capture system (MX System, Vicon Motion Systems, Ltd.) recording the transitions with markers attached at the ankle, knee, pelvis, shoulders, and head of the dummy. The angles measured were computed from the hips to shoulder markers in the coronal for the trunk rotation and sagittal plane for trunk raise. The arm motion was computed from the right shoulder to the right hand marker from the coronal plane.

\begin{figure}[!t]
      \centering
		\subfigure[Motion data during sitting.]{\includegraphics[width=4.2cm]{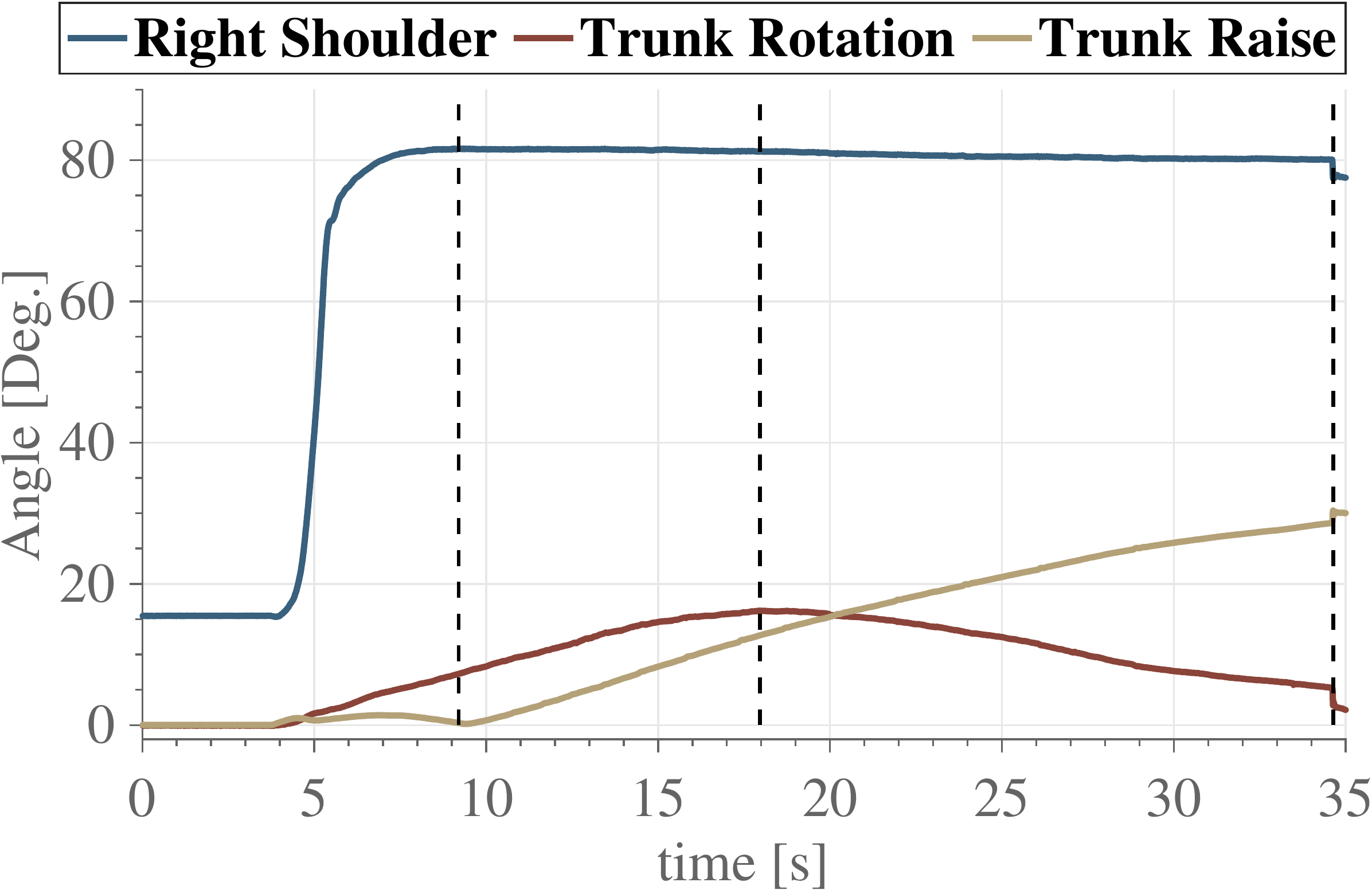}%
		\label{fig:sit_test}}
		\hfil 
		\subfigure[Motion data during lying.]{\includegraphics[width=4.2cm]{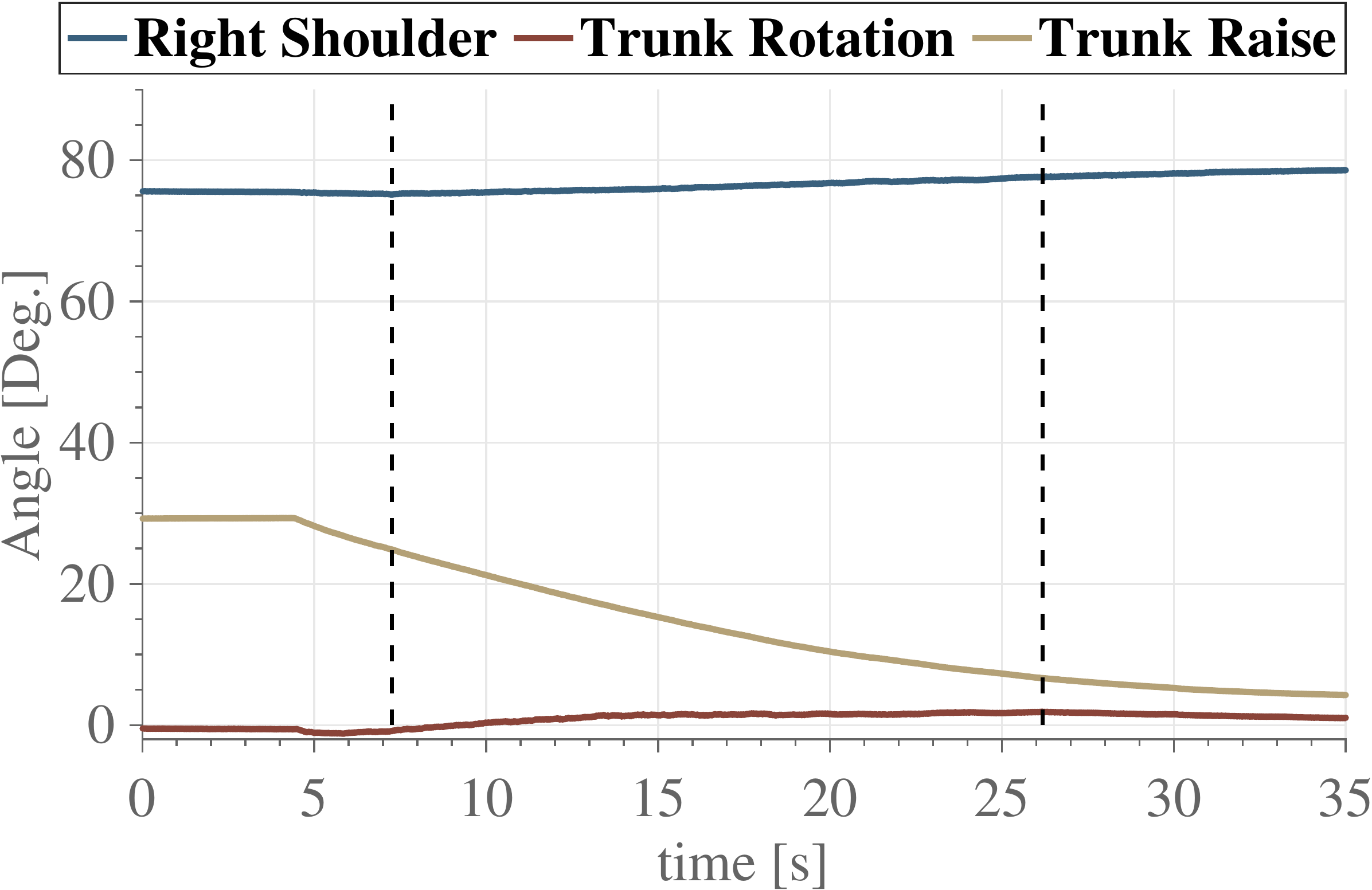}%
		\label{fig:lying_test}}
		\subfigure[SIA pressure during sitting.]{\includegraphics[width=4.2cm]{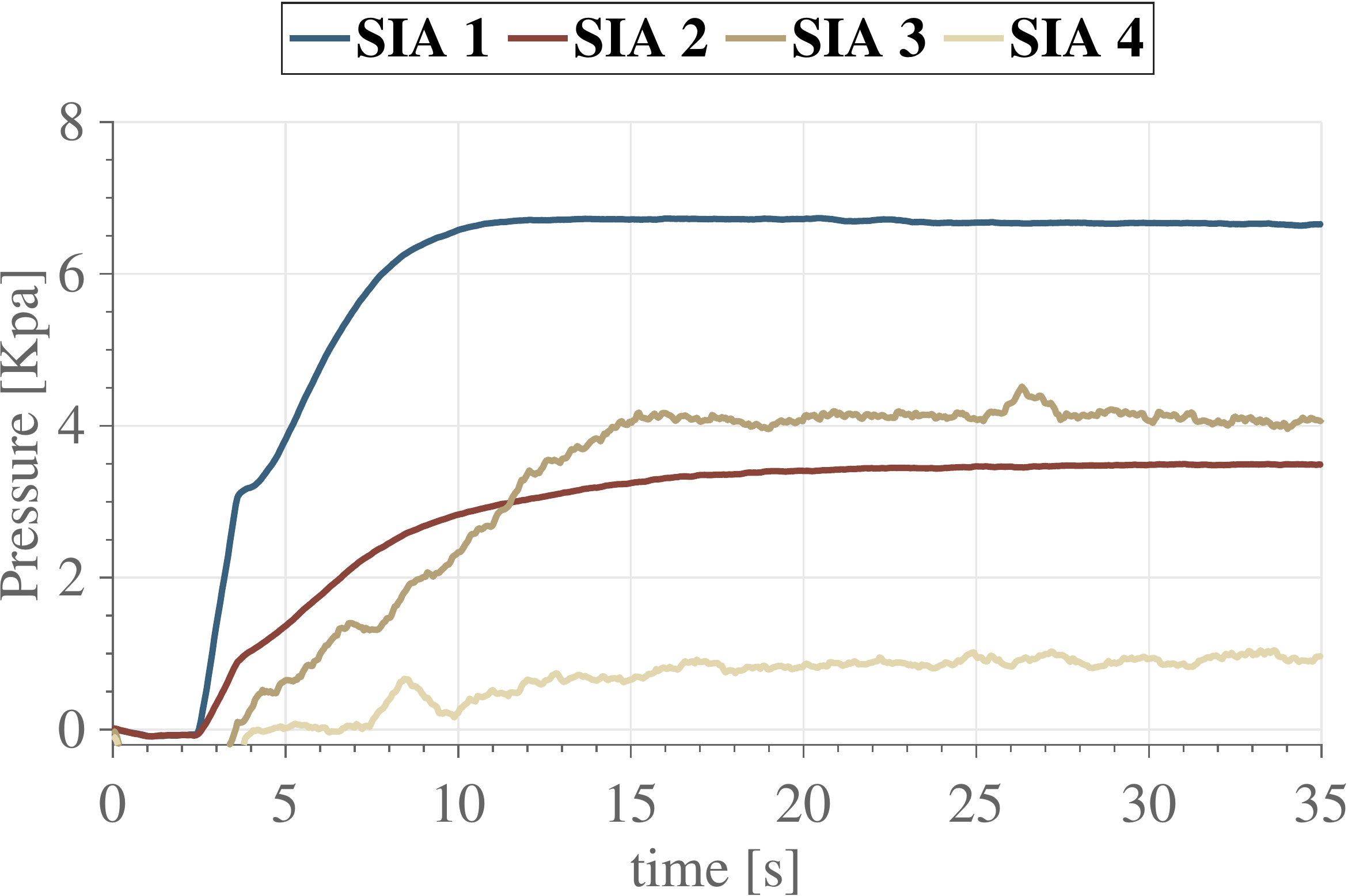}%
		\label{fig:sitting_P}}
		\hfil 
		\hfil 
		\subfigure[SIA pressure during lying.]{\includegraphics[width=4.2cm]{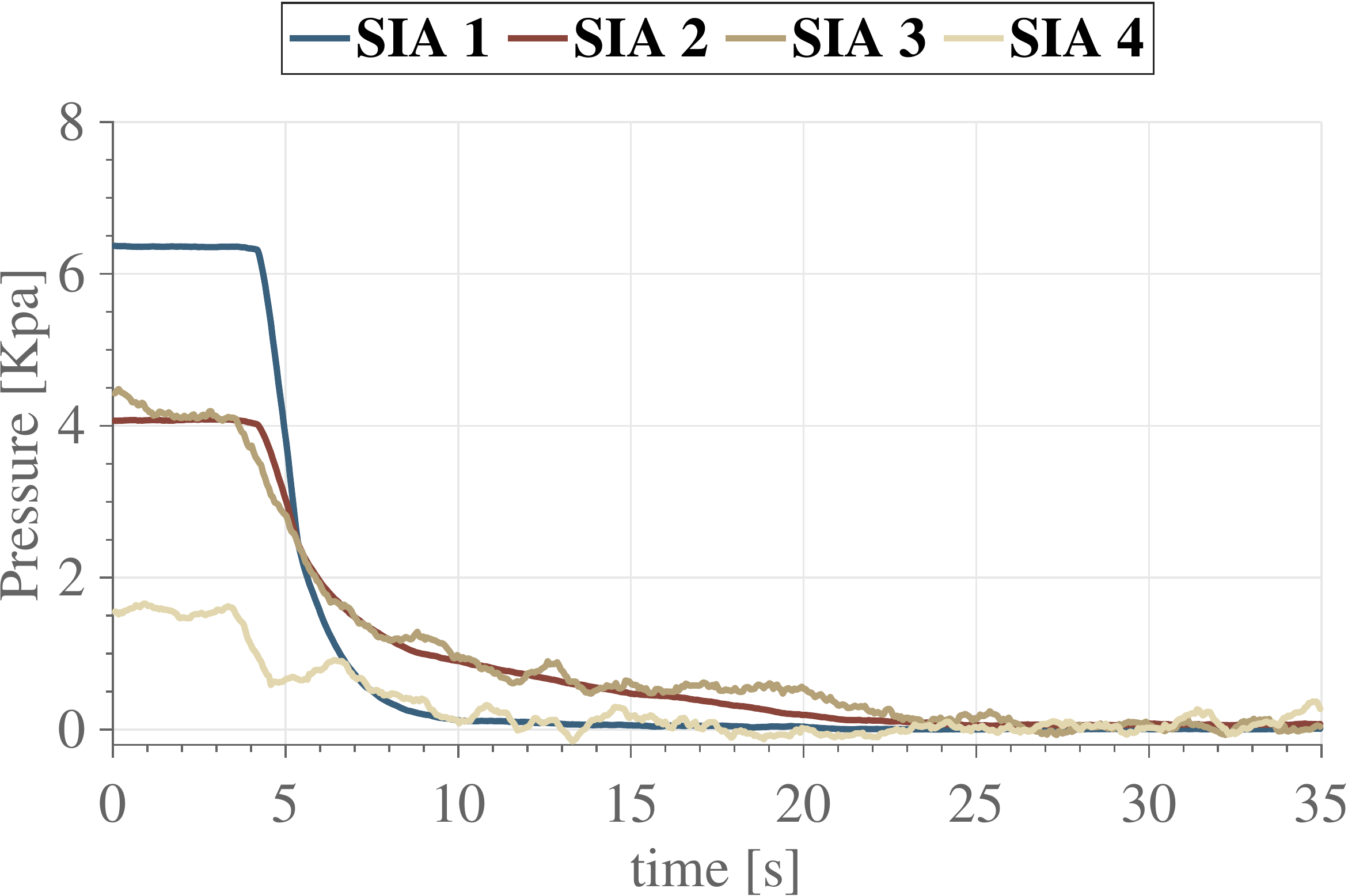}%
		\label{fig:lying_P}}
      \caption{Motion capture data from sitting-lying postural transitions using a dummy baby with $82 cm$ height and $7.95 kg$ weight. 
      \label{fig:test_SIA} }
   \end{figure}

Fig. \ref{fig:sit_test} shows the sequential activation of the robot suit reaching its maximum value in each DOF at  $t=9.2s$ for the right shoulder with $80^o$, $t=17.96 s$ for the trunk rotation with $16.2^o$, and at $t=34.63s$ for the trunk raise with an angle of $34.04^o$.
The pressure reached for each actuator through the series circuit is depicted in Fig. \ref{fig:sitting_P}, where the effect of pressure drop between subsequent SIA shows the same behavior as in Fig. \ref{fig:res_series_cont}. Similarly, sitting transition in  Fig. \ref{fig:lying_P} and \ref{fig:lying_test}, shows a timed motion on the dummy's body although the pressure change in the SIA is very small.

Multiple tests of the wearable soft-robot suit showed the same patterns presented in Fig. \ref{fig:test_SIA}, with differences in the angles based on the posture error of the actuators inside the garment arrangement, some examples are shown in the attached multimedia material.

\section{Discussion and Conclusion}\label{sec:discuss}
The goal of this work was to provide a simple method of controlling the sequential activation of multiple series inflatable actuators (SIA), which was achieved by controlling the pressure through fully passive flow resistors. Several layouts in parallel and series circuits demonstrated the flexibility in designing timed motion. Finally, the robot-suit was accomplished through a series circuit with only one active on/off bypass valve and a pressure regulator. Herewith, simplifying a multi-actuator sequential control.

We strongly believe in the potential applications of this new soft robot-suit for assisting the lying-to-sitting (LtS) and sitting-to-lying (StL) transitions in a clinical or home setting. Especially envisioned for supporting children with cerebral palsy, a disorder that limits muscle control, requiring assistance to move at all times during infancy. Furthermore, other populations could benefit from similar soft supports with multi-degrees of freedom for postural change on a bed, on a wheelchair, or on other wearables for shoulders or neck as rehabilitation support, which is difficult to achieve by motor-based exoskeletons.
Moreover, providing postural assistance to users through a soft physical interface generates a safe environment for both the assisted users and caregivers, allowing them to freely interact with the user's body.

In conclusion, the proposed SIA as a combination of inflatable actuators and passive flow resistors allows sequential control of multiple DOF soft-robots with predefined and coordinated motions. Therefore, we expect many applications to benefit from using the proposed passive flow resistors with any other type of actuators, such as multi-finger actuation in soft grasping, inner pneumatic muscles for soft manipulators, tube-like inflatable actuators, or multi-limb soft robots. e.g., a pneumatic gripper using the proposed flow resistor circuits could have its finger joints engaged before precedent actuators at the hand even if serially connected. 

Further investigation should continue to explore flow control modeling for more precisely timed control activation through the passive flow resistors and integration into more complex actuators.




\bibliographystyle{IEEEtran}
\bibliography{IEEEabrv,bib_RAL_SSIA}





\end{document}